%% file: main.tex
\documentclass{article}

\usepackage[preprint]{corl_2026} 

\usepackage{times}
\usepackage{enumitem}

\hypersetup{
	colorlinks=true,
	linkcolor=orange,
	filecolor=magenta,      
	urlcolor=orange,
	citecolor=magenta,
}
\usepackage{xcolor}
\usepackage{amsmath}
\usepackage{amsfonts}
\usepackage{graphicx}
\usepackage{wrapfig}
\usepackage{booktabs}
\usepackage{multirow}
\usepackage{subcaption}
\usepackage{comment}

\usepackage[capitalise]{cleveref}
\crefname{section}{Sec.}{Secs.}
\Crefname{section}{Sec.}{Secs.}

\crefname{subsection}{Sec.}{Secs.}
\Crefname{subsection}{Sec.}{Secs.}

\crefname{appendix}{App.}{Apps.}
\Crefname{appendix}{App.}{Apps.}

\usepackage{titlesec}

\titlespacing*{\section}{0pt}{0.1em}{0em}
\titlespacing*{\subsection}{0pt}{0.1em}{-0.1em}

\usepackage{caption}
\usepackage{subcaption}
\usepackage{etoolbox}

\usepackage{ifthen}
\newcommand{\todo}[2][none]{%
  \ifthenelse{\equal{#1}{andy}}{%
    \textcolor{blue}{\textbf{TODO:} #2}%
  }{%
  \ifthenelse{\equal{#1}{will}}{%
    \textcolor{orange}{\textbf{TODO:} #2}%
  }{%
  \ifthenelse{\equal{#1}{andrew}}{%
    \textcolor{purple}{\textbf{TODO:} #2}%
  }{%
    \textcolor{red}{\textbf{TODO:} #2}%
  }}}%
}

\newcommand{\forappendix}[1]{}

\newcommand{\loose}{\looseness=-1}
\DeclareMathOperator*{\argmax}{arg\,max}

\newcommand{\methodName}{FRS}

\title{Improving Robotic Generalist Policies via\\Flow Reversal Steering}
\author{
  Andy Tang$^{*, 1}$, William Chen$^{*, 2}$, Andrew Wagenmaker$^{2}$, Chelsea Finn$^{1}$, Sergey Levine$^{2}$\\
  Stanford University$^{1}$, UC Berkeley$^{2}$\\
  \url{flow-reversal-steering.github.io}
}

\begin{document}

\maketitle
\vspace{-3em}
\begin{abstract}
    Generalist policies can learn a wide range of skills from diverse robot datasets. In order to solve or improve on challenging new tasks, we need a way to infer and invoke the appropriate actions from the policy's rich behavioral prior, especially when directly commanding the policy fails. We focus on flow matching generalists and propose Flow Reversal Steering (FRS): a method that takes suboptimal but ``reasonable'' actions, finds their latent noises by passing them through the flow policy in reverse, and maps them to nearby generalist action modes. We evaluate FRS across many simulated and real-world manipulation settings. First, FRS can turn coarse semantic guidance from humans or vision-language models (VLMs) into corresponding good robot actions, improving zero-shot control. These gains can be distilled with behavioral cloning by training an auxiliary policy to output noises that the generalist maps to good actions -- showing up to 95\% absolute task success rate boosts in under a minute of training. Finally, FRS enables policy improvement by bootstrapping reinforcement learning with semantic knowledge, improving on several tasks that standard RL fails to improve on.
\end{abstract}

\input{sections/01-introduction}

\input{sections/02-related_works}
\input{sections/03-preliminaries}

\input{sections/04-method}

\input{sections/05-experiments}
\input{sections/06-discussion}

\clearpage
\acknowledgments{We would like to thank Qiyang Li, Seohong Park, Charles Xu, Jubayer Ibn Hamid, Alex Swerdlow, Lars Ankile, Arhan Jain, Jenny Pan, Ayush Agrawal, Jesse Zhang, and the other members of RAIL and IRIS Labs for insightful discussions and help with experiments.
This research was partly supported by ONR N00014-25-1-2060, ARL DCIST CRA W911NF-17-2-0181, and DARPA TIAMAT.
We would like to thank the NVIDIA Academic Grant Program for providing compute resources.
This work used GPUs at NCSA Delta and Purdue Anvil through allocation CIS260400 from the Advanced Cyberinfrastructure Coordination Ecosystem: Services \& Support (ACCESS) program, which is supported by U.S. National Science Foundation grants \#2138259, \#2138286, \#2138307, \#2137603, and \#2138296 \cite{accessCompute}.}

\bibliography{refs} 

\newpage
\input{sections/appendices}
\end{document}

%% file: sections/01-introduction.tex
\vspace{-0.5em}
\begin{figure}[h]
    \centering
  \includegraphics[width=\textwidth]{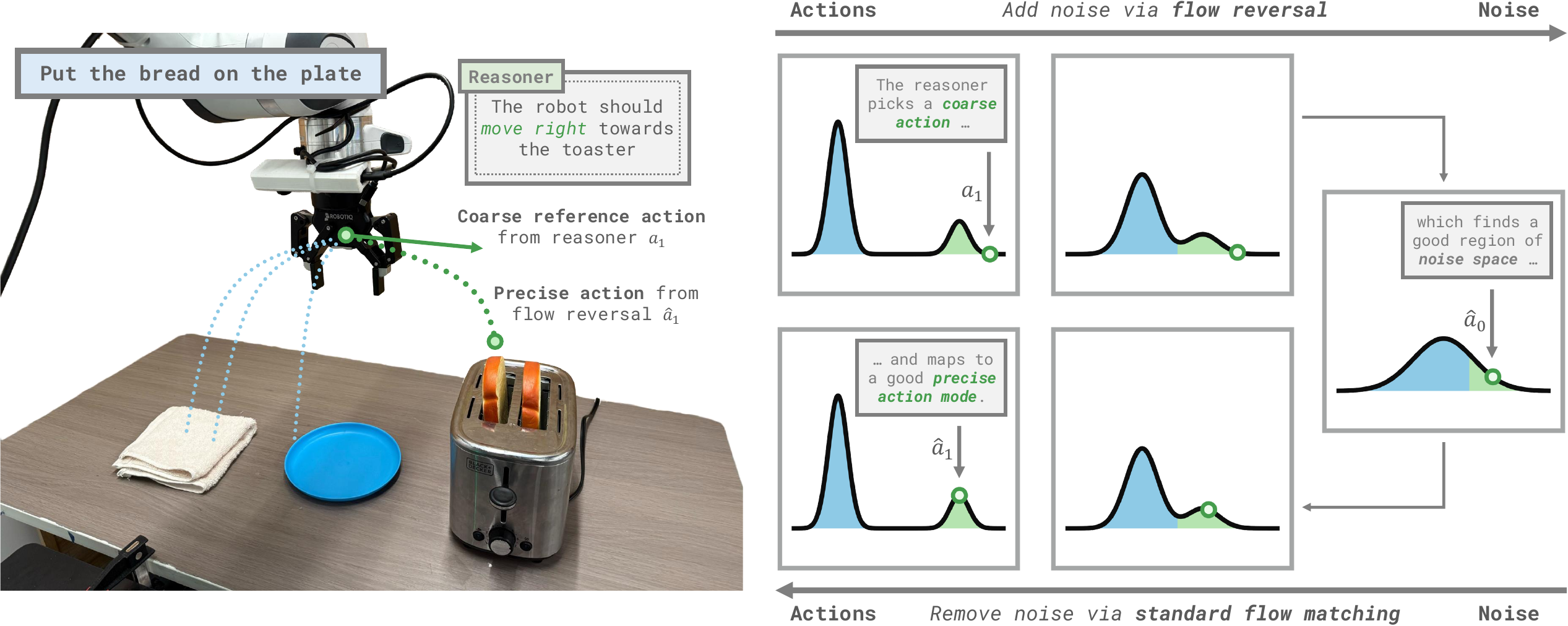}
  \vspace{-1.5em}
  \captionof{figure}{
    Flow Reversal Steering (FRS) uses semantic reasonings from humans or VLMs to steer generalist flow policies towards sampling beneficial action modes. It does this by passing coarse but ``reasonable'' actions through the flow model \textit{in reverse} to identify good regions of noise space to denoise. This enables effective task learning and adaptation via noise-space behavioral cloning and reinforcement learning.
  }
  \label{fig:teaser}
  \vspace{-0.5em}
\end{figure}

\section{Introduction}
\label{sec:intro}

Robotic foundation models trained on large and diverse datasets provide a powerful recipe for learning multi-task generalist policies~\citep{kim2024openVLA, brohan2023rt2, tri_lbm2025lbm, ye2026wam-worldActionModels, pai2025mimicvideo-videoActionModels}.
While such policies can often follow many commands, they will inevitably encounter new tasks diverging from their training data that require longer-horizon behaviors or demand adaptation through test-time trial-and-error. The standard recourse in such situations would be to simply add more demonstration data, retrain the generalist, and try again. However, we observe that the knowledge in these models goes beyond simply following instructions -- it provides a rich prior over reasonable behaviors. For example, a policy trained to interact with bowls, sponges, and towels has many of the skills needed when learning new kitchen-cleaning tasks, like wiping countertops or cleaning dishes. Effectively invoking appropriate actions from this prior would allow for rapid adaptation. The question then becomes: how do we best access the prior knowledge in generalist policies when faced with new tasks?

\textit{Policy steering} -- that is, guiding the action-sampling process to direct policy outputs to some desired end -- offers a way to use the generalist's ``reasonable'' action prior in novel tasks by upweighting relevant behaviors. 
In particular, steering could allow generalists to make use of knowledge from \textit{semantic reasoners}, such as humans and large vision-language models (VLMs). For instance, when the policy is learning the novel task of cleaning a kitchen countertop, the knowledge that ``sponges are used for wiping spills'' could be used to steer the policy to reach for the sponge, instead of attempting other behaviors the robot could reasonably do in the scene. We thus want a steering method that is suitable for eliciting good actions from the generalist prior, based on semantic inferences.\loose

Many steering methods use diffusion or flow matching~\citep{ho2022cfg, dhariwal2021diffusion-classifier-guidance, singhal2025feynmanKacSteering, yoneda2025toNoiseAndBack, wang2024inference-time-policy-steering, frans2025cfg-rl, ho2020ddpm-denoisingDiffusionProbabilisticModels, chi2024diffusionPolicy}, a common parameterization for generalist policies. Notably, flow matching policies learn a \textit{deterministic} map from noise to action~\citep{lipman2023flowMatching} that can be steered by finding the noise values that map to desirable actions within the generalist prior~\citep{wagenmaker2025dsrl}. However, the noise space of flow policies lacks immediately-apparent structure, so past works resort to expensive trial-and-error via reinforcement learning to find good noise values. To effectively steer generalist flow policies, we would instead ideally utilize semantic reasoning to quickly identify noise values that map to semantically-appropriate actions.

We thus propose \textbf{Flow Reversal Steering (\methodName{})}: a novel approach that maps coarse reference actions to their noises by passing them through flow policies \textit{in reverse}. When denoised, this yields actions that are fine-grained and ``in-distribution'' for the generalist, while staying roughly consistent with the reference action. Given even a rough sketch of robot behaviors (e.g., the general direction needed to reach for a target object), \methodName{} can ``project'' that behavior into the generalist's prior to produce a similar fine-grained action. 

This mechanism is especially useful for semantic reasoners, such as VLMs, that can roughly infer appropriate robot behaviors, but cannot ground them into dexterous low-level actions. \methodName{} moves the onus of emitting robot actions to the generalist, while reasoners can focus on broad, high-level steering.
As \methodName{} also gives corresponding noise vectors, it meshes well with latent-noise policies, which steer flow generalists by changing their distribution of input noises. This can both (1) enable fast and efficient adaptation via noise-space behavioral cloning (BC) and (2) bootstrap noise-space reinforcement learning (RL) for tasks where exploring via the generalist policy is intractable.

We evaluate \methodName{} on state-of-the-art generalist vision-language-action policies (VLAs~\citep{pi2025pi05}) through extensive simulated and real-world manipulation experiments~\citep{liu2023libero, khazatsky2024droid}. We show how \methodName{} allows humans and VLMs to effectively guide generalists across diverse tasks, even by simply specifying just the rough direction the robot should move. Then, we show how the noises from \methodName{} can be used for robustly learning tasks in just one minute of active BC training on 10 trajectories. Lastly, we find \methodName{} speeds up noise policy RL, where using just one \methodName{} success as prior data enables efficient improvement on tasks that the base VLA nearly always fails at.

%% file: sections/02-related_works.tex
\section{Related Works}
\label{sec:related_works}

\textbf{Foundation models for robotics}. Large pretrained foundation models~\citep{bommasani2022foundationModels} can be applied in robotics in two ways: by fine-tuning them on robotic data \textit{or} by querying them zero-shot. The former can involve training them with BC on large-scale demonstration datasets~\citep{embodimentcollaboration2024oxe, khazatsky2024droid, jiang2025galaxeaOpenWorldDatasetG0}, with two popular approaches fine-tuning (1) VLMs into vision-language-action models~\citep{brohan2023rt2, kim2024openVLA, black2024pi0} or (2) video generators into world-action models~\citep{pai2025mimicvideo-videoActionModels, ye2026wam-worldActionModels}. These policies are exposed to many robotic skills during pretraining, letting them follow many user commands. Vitally, this also captures a prior distribution over ``reasonable'' behaviors, often including behaviors needed to solve novel tasks the policies initially fail at. \methodName{} aims to flexibly elicit behaviors from this prior that are semantically appropriate for novel tasks, which in turn can be used for rapid adaptation and policy improvement. VLMs can also be trained into reward models for RL~\citep{liang2026robometer, lee2026roboReward, sontakke2023roboclip, zhai2025vlac-visionlanguageactioncriticmodel}. This is orthogonal and complementary to our work, as using \methodName{} for RL places no restrictions on the chosen reward function.

Both actions and rewards can also be predicted by foundation models \textit{without} training on robot data. VLMs can both invoke robot behaviors via predefined interfaces~\citep{liang2023codeAsPolicies, singh2022progPrompt, ha2023scalingUpDistillingDown, vemprala2023chatgptRobotics, shi2025maestro, fu2026capX, huang2023voxPoser, kumar2026vlmTAMP, shen2026tiptop, nasiriany2024pivot, liu2024moka, sathyamoorthy2024convoi} or act as reward functions~\citep{ma2024gvl-incontextValueVLM, rocamonde2024visionlanguagemodelszeroshotreward, chen2026topreward, budzianowski2026opengvl, zhang2026progresslm, ma2024eureka}. However, such methods are limited by VLMs' capabilities. While VLMs can effectively compose a limited set of high-level behavioral primitives, they struggle when given lower-level ones, thus giving a tradeoff between performance and flexibility~\citep{fu2026capX}. VLMs also struggle with reward modeling due to their limited fine-grained visual reasoning skills~\citep{zhang2026progresslm}. Our method avoids this issue by relying on the generalist to produce actions. Instead, VLMs can make the high-level semantic inferences they excel at, while \methodName{} grounds this coarse feedback into appropriate actions.

\textbf{Diffusion policies and steering.} Many generalist policies use diffusion or flow matching~\citep{ho2020ddpm-denoisingDiffusionProbabilisticModels, lipman2023flowMatching, chi2024diffusionPolicy, tri_lbm2025lbm, black2024pi0, pi2025pi05, bjorck2025gr00t, ye2026wam-worldActionModels}, which iteratively denoises Gaussian noise vectors to generate action samples. 
They can be steered by either modifying this denoising process or changing the partially-noised data inputs that get denoised~\citep{ho2022cfg, dhariwal2021diffusion-classifier-guidance, singhal2025feynmanKacSteering, pi2025pi06star, frans2025cfg-rl, yoneda2025toNoiseAndBack, wang2024inference-time-policy-steering}. Notably, flow matching \textit{deterministically} maps noise to outputs, so it can be steered by finding pure noises that denoise to good actions. 
However, finding good noises is challenging -- they are usually identified by trial-and-error with RL~\citep{wagenmaker2025dsrl}. In contrast, we propose combining flow reversal with coarse feedback to quickly identify effective noise. We show how flow reversal enables efficiently training policies that emit noises to steer the generalist toward solving new tasks, \textit{while obviating or accelerating the tedious process of discovering good noises via RL.}

\textbf{Improving generalist policies.} Past methods for improving generalists (like VLAs) with RL often have certain traits. 
When tuning the full VLA, RL methods tend to use supervised learning -- like distillation -- for policy extraction~\citep{xu2024rldg-roboticGeneralistPolicyDistillation, mark2024parl-policyAgnosticRL, pi2025pi06star,  xiao2025pld-probeLearnDistill}, avoiding needing action probabilities (which are hard to extract from flow models). They are often batched online, as their size makes true online RL unwieldy.
To avoid this issue, other works train separate smaller policies with online RL~\citep{xu2024rldg-roboticGeneralistPolicyDistillation, xiao2025pld-probeLearnDistill, wagenmaker2025dsrl}, often using the base policy prior to constrain behavior to ``reasonable'' actions, e.g., via residual RL~\citep{xiao2025pld-probeLearnDistill, johannink2018residualRL}, behavior attenuation with classifier-free guidance~\citep{pi2025pi06star, frans2025cfg-rl}, or treating the VLA as a latent action decoder~\citep{wagenmaker2025dsrl}. These policies can then be distilled into the generalist~\citep{xiao2025pld-probeLearnDistill, xu2024rldg-roboticGeneralistPolicyDistillation}.
Our method uses semantic feedback to elicit ``reasonable'' behaviors during generalist policy improvement, yielding gains beyond existing work which solely use the base policy as a behavior constraint.

\textbf{Flow and diffusion reversal in robotics.}
While flow or diffusion reversal is common in vision (\Cref{app:related_additional}), fewer works use such techniques in robotics. GenPO~\citep{ding2026genpo} inverts actions from a diffusion policy to estimate their likelihood for RL updates. Concurrent to our work, UniSteer~\citep{lu2026unified} inverts human actions via flow reversal to obtain ``good'' noises, which are used for noise-space RL by adding a behavioral cloning (BC) term. 
While similar, our approach diverges from these works in three key ways: (1) we use flow reversal to steer generalist flow policies by \textit{refining} coarse actions, admitting guidance from humans \textit{and} scalable VLMs; (2) we show how, even without RL, flow reversal enables efficiently training noise policies with \textit{just} BC; and (3) we show that flow reversal can bootstrap RL with coarse non-human guidance, even if the base policy nearly never succeeds.

%% file: sections/03-preliminaries.tex
\section{Preliminaries}
\label{sec:prelim}

\textbf{Flow matching VLAs.} Generalist policies take a pretrained backbone and fine-tunes it with behavioral cloning (BC) to match the action distribution of experts. Given BC data $(o, a_1)$ and sampled times $t \in [0, 1]$, these policies fit a velocity field $v_\theta(a_t, t \mid o)$ that maps noise $a_0 \sim \mathcal{N}(0, I)$ to actions $a_1$ by denoising partially-noised actions produced by forward diffusion $a_t = t \cdot a_1 + (1 - t) \cdot a_0$.
The flow $v_\theta$ satisfies an ordinary differential equation by minimizing the loss $\mathcal{L}_\theta$:
\begin{equation*}
    \mathcal{L}_\theta = \mathbb{E}_{o, a_1, a_0, t} \left[ \left|\left| v_\theta(a_t, t \mid o) - (a_1 - a_0) \right|\right|^2 \right] \Longleftrightarrow \text{d}a_t = v_\theta(a_t, t \mid o) \ \text{d}t
\end{equation*}
$a_1$ can be practically solved for via Euler integration with time steps $h$ (with $1/h$ iterations):
\begin{equation*}
a_{t+h} \leftarrow a_t + v_\theta(a_t, t \mid o) \cdot h, \quad\quad \text{for } t \in \left\{ 0, h, \dots, 1 - h \right\}, \quad a_0 \sim \mathcal{N}(0, I)
\end{equation*}
We denote this denoising process as $a_1 \leftarrow \mu_\theta(a_0, o)$. This defines the BC policy, which samples actions by drawing $a_0 \sim \mathcal{N}(0, I)$ and denoising it with $\mu_\theta$ into $a_1$. We specifically consider flow matching VLAs~\citep{black2024pi0, pi2025pi05, bjorck2025gr00t}, though in principle, our method is applicable to other flow matching generalist architectures or diffusion policies with deterministic DDIM sampling too~\cite{chi2024diffusionPolicy, song2020denoising}.

\textbf{Diffusion steering.} Flow policies' outputs can be controlled by finding noises that map to desirable actions, rather than drawing from $\mathcal{N}(0, I)$. One way is to partially noise a reference action, then denoise it with the BC policy $\mu_\theta$ to produce a similar one~\citep{wang2024inference-time-policy-steering, yoneda2025toNoiseAndBack}. This fails when fully noising, as doing so removes all information about the reference. Good noises can also be found via RL; Diffusion Steering via Reinforcement Learning (DSRL~\citep{wagenmaker2025dsrl}) learns an auxiliary \textit{latent noise action policy} $\pi_\phi^\text{noise}(a_0 \mid o)$ by treating the noise $a_0$ as an action and running RL in a ``noise action'' Markov decision process. It learns a noise critic $Q^{\text{noise}}(o, a_0)$, and trains the actor $\pi_\phi^\text{noise}(a_0 \mid o)$ to maximize it:\loose
\begin{equation*}
\textstyle \pi_\phi^\text{noise} \leftarrow \argmax_\pi \mathbb{E}_{o \sim \mathfrak{B},a_0 \sim \pi(\cdot \mid o)}[Q^{\text{noise}}(o, a_0)]
\end{equation*}
for replay buffer $\mathfrak{B}$. At deployment, DSRL samples $a_0 \sim \pi_\phi^\text{noise}(\cdot \mid o)$, then computes and executes $a_1 \leftarrow \mu_{\theta}(a_0, o)$. $\pi_\phi^\text{noise}$ thus finds noises that the BC policy $\mu_\theta$ denoises to good robot actions.

%% file: sections/04-method.tex
\section{Flow Reversal Steering (\methodName{})}
\label{sec:method}

\begin{figure}[t]
    \centering
    \includegraphics[width=\linewidth]{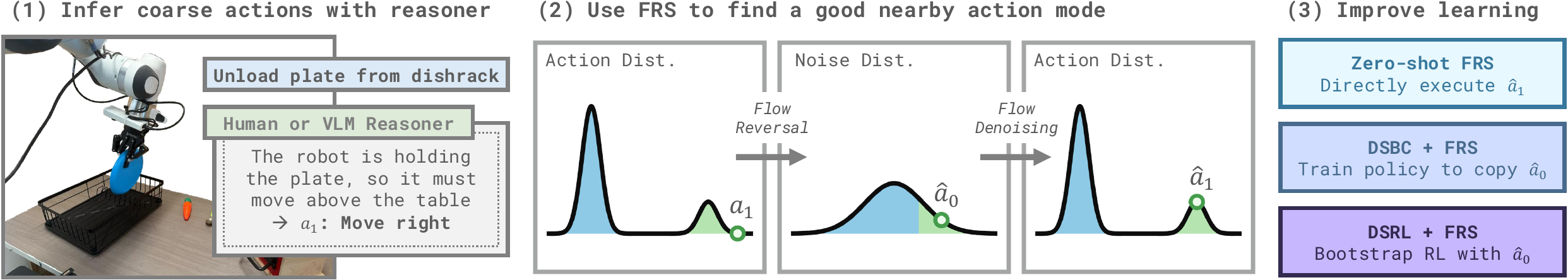}
    \caption{Overview of FRS. \textbf{(1)} A human or VLM semantically reasons about the novel task to determine a reference action capturing roughly what the robot should do. \textbf{(2)} This coarse action is passed through flow reversal and denoising, projecting it into the space of generalist actions. \textbf{(3)} Both the expert noises and actions can be used for policy improvement by executing the action (zero-shot \methodName{}, \cref{subsec:zero-shot-experiments}) or training a noise action policy with supervised learning (DSBC, \cref{subsec:dsbc-experiments}) or reinforcement learning (DSRL + \methodName{}, \cref{subsec:rl-experiments}).}
    \label{fig:frs-overview}
    \vspace{0.2cm}
    \centering
    \includegraphics[width=\linewidth]{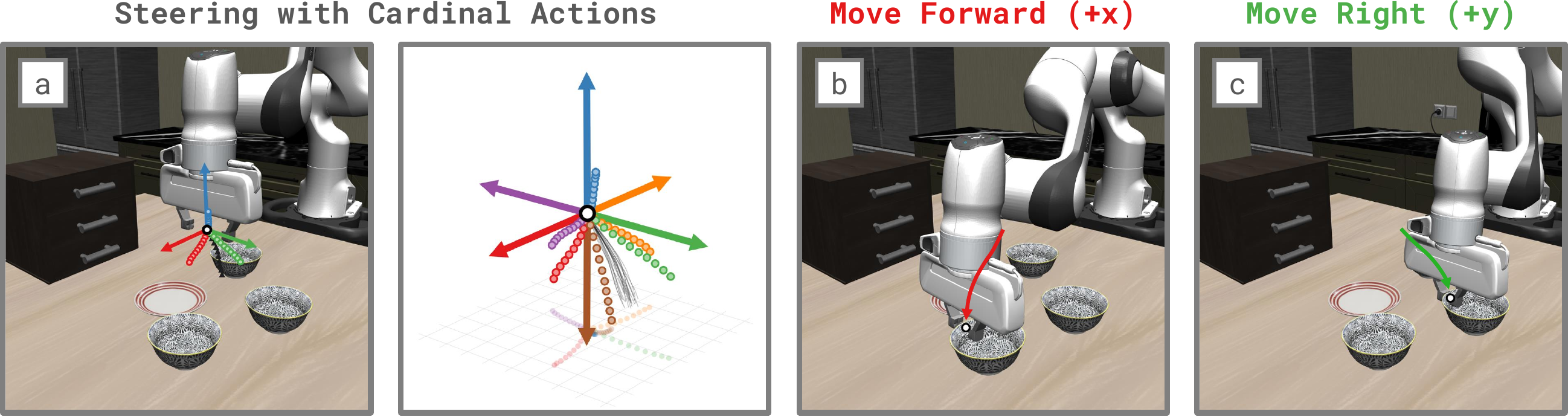}
    \caption{Illustrative examples of \methodName{} with $\pi_{0.5}$ in LIBERO. \textbf{(a)} Solid arrows are directional reference actions, dots are corresponding steered actions, and black represents samples from the base policy without steering. The outputs of \methodName{} are biased towards ``reasonable'' behaviors given the scene, e.g., reaching for the bowls. \textbf{(b)} and \textbf{(c)} show rolling out the forward (red) and right (green) steered actions, followed by executing the base policy.
    }
    \vspace{-1em}
    \label{fig:main-frs-example}
\end{figure}

Our method, Flow Reversal Steering (FRS), takes ``coarse'' actions and refines them using a generalist policy into similar, higher-quality actions. 
In \cref{subsec:flow-reversal}, we show how flow reversal can identify noises that bias the flow matching policy into sampling actions \textit{of the same mode} as the coarse one.
Then, in \cref{subsec:vlm-priors}, we consider how humans or VLMs can provide such coarse guidance to the robot, based on their semantic knowledge. Flow reversal converts these rough sketches of robot behaviors into in-distribution actions.
Finally, in \cref{subsec:steering-improvement}, we present ways to use these semantically-guided trajectories to efficiently learn and improve at new tasks.
See \cref{fig:frs-overview} for an illustrative overview.

\subsection{Reversing the Flow Velocity Field}
\label{subsec:flow-reversal}

Since flow denoising is deterministic for a given input, flow can be \textit{reversed} to determine the noise $a_0$ that corresponds to a given reference action $a_1$ by simply integrating the ODE it defines backwards in time. That is, using Euler integration:
\begin{equation*}
    a_{t-h} \leftarrow a_t - v_\theta(a_t, t \mid o) \cdot h, \quad \text{for } t \in \left\{ 1, 1-h, \dots, h \right\}
\end{equation*}
We denote this \textit{noising} process as $\hat{a}_0 \leftarrow \mu_\theta^{-1}(a_1, o)$, as it inverts $\mu_\theta$, where the hat denotes a \textit{computed} noise, not a sampled one. This does not modify the learned model $v_\theta$ at all -- flow reversal needs the same computations as standard flow denoising, just starting from an action instead of noise and swapping the order of integration. In turn, $\hat{a}_0$ can be passed through standard flow denoising to yield $\hat{a}_1 \leftarrow \mu_\theta(\hat{a}_0, o)$. As $h \rightarrow 0$, this exactly reconstructs the reference $\hat{a}_1 = a_1$. However, for finite iterations, integration error means that $\hat{a}_1$ only approximately reconstructs $a_1$.

Empirically, we find that this results in reconstructed actions $\hat{a}_1$ being \textit{similar but not identical to} the reference $a_1$, while also being ``in-distribution'' for BC policy $\pi_\theta$. This yields actions from the generalist that are \textit{biased} towards reference behaviors, rather than perfectly reconstructing them. We call this \textbf{Flow Reversal Steering (FRS)}.

We show an illustrative example of \methodName{} with an actual VLA in \cref{fig:main-frs-example} (see \cref{app:illustrative-flow-reversal-expts} and \cref{fig:libero-steering-examples} for more), using ten noising/denoising steps ($h=0.1$). 
The steered actions follow the same general direction as their respective coarse reference, albeit biased by the affordances in the scene -- e.g., when the gripper is above the table and empty, the steered actions tend to move down towards objects to grasp; 
\begin{wrapfigure}[16]{r}{0.5\linewidth}
  \centering
  \vspace{-1.1em}
  \includegraphics[width=\linewidth]{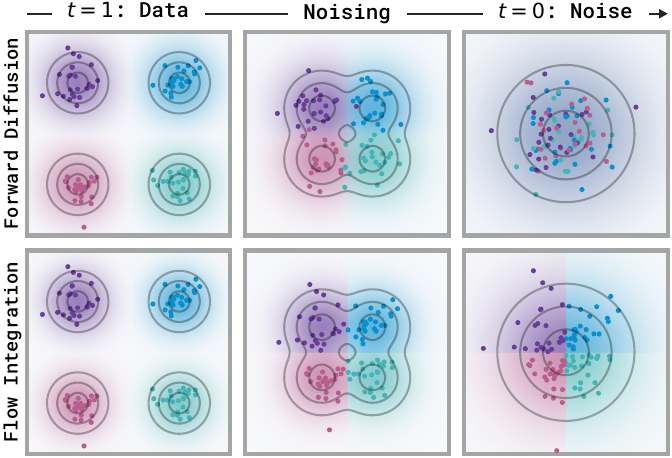}
  \caption{Noising via the forward diffusion process vs. reverse flow integration. Both have the same marginals, but the former uses noise interpolation (so all signal is gone by $t=0$), while the latter deterministically maps from data to noise and back.}
  \label{fig:flow-vs-forward-diffusion}
  \vspace{-0.2cm}
\end{wrapfigure}
when the gripper is holding something, it tend to move up to lift the object \textit{or} towards containers to place it. These are the ``reasonable'' actions internalized by the generalist policy.

Note that flow reversal is \textit{distinct} from the forward diffusion process, which linearly interpolates data with Gaussian noise $a_t = t \cdot a_1 + (1 - t) \cdot a_0;\ a_0 \sim \mathcal{N}(0, I)$. While this is used for training flow models, it rapidly destroys the information in the output, whereas flow reversal identifies the noise which deterministically maps to the reference action (\cref{fig:flow-vs-forward-diffusion}). Past works propose using forward diffusion for steering by partially noising reference actions, then passing them through denoising~\citep{wang2024inference-time-policy-steering, yoneda2025toNoiseAndBack}. However, we find these methods to be highly sensitive to how much noise is added, and thus hard to tune and ineffective (\cref{subsec:zero-shot-experiments}).

\subsection{High-Level Semantic Reasoning for Steering}
\label{subsec:vlm-priors}

We now consider sources of semantically-reasonable reference actions as inputs to \methodName{}. 
Naturally, humans can guide robots to solve tasks, though standard methods, like teleoperation, are costly and tedious (though can also be used with flow reversal, see \cref{subsec:steering-improvement}).
Similarly, VLMs can roughly identify appropriate robot high-level behaviors~\citep{liu2024moka, nasiriany2024pivot}, even if they cannot emit fine robot actions.
Thus, we need a way for these \textit{reasoners} to tap into their semantic knowledge and easily ground it in coarse actions for guiding the robot.

When running \methodName{} online, we opt to have both human and VLM reasoners emit simple \textit{directional} actions to guide the robot.
That is, the reasoner can choose Cartesian directions based on how they think the manipulator's end effector should move (see \cref{app:steering}). This is programmatically turned into a rough steering action chunk~\citep{zhao2023aloha-act} that servos the robot straight in the specified direction.
Unsurprisingly, such action chunks are ineffective when directly executed (\cref{subsec:zero-shot-experiments}), but are nonetheless suitable as reference actions $a_1$ for steering.
Finally, both reasoners also have the option to defer to the base policy when steering is inappropriate, e.g., when executing precise grasps.

The online \methodName{} inference loop thus involves (1) querying the human or VLM reasoner at each step to infer the general direction the robot should move; (2) converting that motion into a corresponding directional reference action $a_1$; (3) using flow reversal to map it back to noise $\hat{a}_0 \leftarrow \mu^{-1}_\theta(a_1, o)$; and (4) denoising it back into an action to execute $\hat{a}_1 \leftarrow \mu_\theta(\hat{a}_0, o)$. This steers the generalist's action generation based on what the high-level reasoner infers to be useful for the task.

\subsection{Improving Generalist Policies with \methodName{}}
\label{subsec:steering-improvement}

Now that the reasoner can guide the robot toward semantically-sensible behaviors, how can this be used to improve performance? We propose three paradigms for using \methodName{} to improve policies.

\textbf{Zero-shot online steering.} The simplest way is to use \methodName{} \textbf{zero-shot}, having the reasoner steer the policy every step. That is, each time the generalist policy would be queried during deployment, the human or VLM reasoner is queried to produce a semantically-meaningful coarse reference action, which is passed through flow reversal and denoising before being executed.

However, constantly querying the reasoner can become expensive (especially for human reasoners), so we also aim to use \methodName{}'s elicited trajectories for learning. While these rollouts can work with any policy learning algorithm, it synergizes especially well with \textit{noise policy} learning (\cref{sec:prelim}). \methodName{} yields noises immediately aligned with the coarse inferences of semantic reasoners, alleviating the usual difficulty of finding good noises with random trial-and-error when training noise policies via RL. This advantage enables two novel methods for efficient generalist policy learning with \methodName{}.

\textbf{Supervised learning on \methodName{} noise actions.} We can treat flow reversal noises as ``expert'' noise actions for supervised learning, rather than RL. Given observation-noise pairs $(o, \hat{a}_0)$ run BC with:
\begin{equation*}
  \textstyle      \pi_\phi^\text{noise} \leftarrow \argmax_{\pi} \mathbb{E}_{o,a_0}[\log \pi(a_0 \mid o)],
\end{equation*}
thereby distilling \methodName{}'s good noise actions into $\pi_\phi^\text{noise}(\hat{a}_0 \mid o)$. At test time, it is treated exactly like a DSRL noise policy, inferring noises $\hat{a}_0$ at each step that get mapped to actions by the generalist $\hat{a}_1 \leftarrow \mu_\theta(\hat{a}_0, o)$. Naturally, we call this \textbf{Diffusion Steering via Behavioral Cloning (DSBC)}.

There are two ways to acquire the noises for supervising DSBC. First, we can use noises from \textbf{online} rollouts of zero-shot \methodName{} collected as described above. As these noises have been denoised and executed, they are verified as mapping to good actions if they lead to task success.
Second, DSBC can also be applied to \textit{existing} robotic demonstrations. Given an observation $o$ and corresponding demonstrator action $a_1$, flow reversal can augment each frame with noise $\hat{a}_0 \leftarrow \mu^{-1}_\theta(a_1, o)$ approximately mapping to $a_1$,
providing entirely \textbf{offline} data for DSBC. The lack of online execution yields a practical trade-off: as flow reversal does not perfectly reconstruct reference actions, offline DSBC does not ensure that reconstructed actions are free from suboptimality or error.
However, this also permits precise reference actions from \textit{any} suitable teleoperation or control interface.

\textbf{Bootstrapping RL with \methodName{}.} When used \textit{in conjunction with} DSRL, \methodName{}'s noise trajectories can be used as prior data and behavior regularization via two simple changes. First, we augment the policy learning loss by adding an auxiliary DSBC loss over successful \methodName{} rollouts, $\mathfrak{D}^+$. That is, rather than simply training $\pi_\phi^\text{noise}$ to maximize $Q^{\text{noise}}$ as standard DSRL does, we train $\pi_\phi^\text{noise}$ via:
{\setlength{\abovedisplayskip}{2pt}
 \setlength{\belowdisplayskip}{15pt}
 \setlength{\abovedisplayshortskip}{1pt}
 \setlength{\belowdisplayshortskip}{2pt}
\begin{align*}
\pi_\phi^{\text{noise}} \leftarrow \arg\max_\pi
\smash{\underbrace{
\mathbb{E}_{o \sim \mathfrak{B},\, a_0 \sim \pi(\cdot \mid o)}
\!\left[ Q^{\text{noise}}(o,a_0) \right]
}_{\text{RL objective}}}
+ \lambda\,
\smash{\underbrace{
\mathbb{E}_{(o,a_0)\sim\mathfrak{D}^{+}}
\!\left[ \log \pi(a_0\mid o) \right]
}_{\text{BC auxiliary objective}}}
\rule[-2ex]{0pt}{0pt}
\end{align*}
}Second, we prefill DSRL's buffer $\mathfrak{B}$ with \methodName{} trajectories (optionally including failed ones).
This enables improvement from experience beyond zero-shot \methodName{} and DSBC. 
We call this \textbf{DSRL + \methodName{}}. 

These changes improve the efficiency of RL by encouraging the noise policy to explore around \methodName{}'s semantically-meaningful behaviors, contrasting the random noise sampling early in DSRL. Furthermore, since robot trajectories do not usually have noise actions, training noise policies with prior data is usually challenging or expensive and requires, for example, distilling robot action space Q-functions into noise action space~\citep{wagenmaker2025dsrl}. Flow reversal circumvents this by rapidly and cheaply identifying good underlying noises from reference actions, including from offline data.

%% file: sections/05-experiments.tex
\section{Experiments}
\label{sec:experiments}

We now evaluate Flow Reversal Steering by answering: 
(1) Can \methodName{} improve performance without any training by having VLMs guide generalists towards semantically-reasonable behaviors? 
(2) Can we use the improved trajectories from \methodName{} to rapidly learn new tasks?
(3) Can \methodName{} help generalists more efficiently \textit{improve} from experience?
See \cref{app:sim} and \cref{app:real-world} for more experimental details.

\subsection{Experimental Setup}
\label{subsec:experimental-setup}

\textbf{Simulation.} We use LIBERO~\citep{liu2023libero} for scalable simulated evaluations. Our zero-shot results consider the full Spatial, Object, and Goal splits, as well as all 62 tasks in 90 that our base policy achieves $\leq$ 40\% success on (\cref{subsec:zero-shot-experiments}). We then use a 15-task subset of LIBERO-90 where \methodName{} achieves sufficient success to train DSBC policies (\cref{subsec:dsbc-experiments}). Finally, we run DSRL + \methodName{} on that subset, as well as a harder 10-task subset where the base policy nearly completely fails (\cref{subsec:rl-experiments}).
To allow room for improvement, we use base VLAs that have \textit{not} been trained on the LIBERO splits that we run them on, following \citet{wagenmaker2025dsrl}. For LIBERO-90, we use OpenPi's $\pi_{0.5}$-LIBERO~\citep{pi2025pi05}, which is trained on all splits \textit{except} 90. For all others, we use $\pi_{0.5}$ fine-tuned by \citet{jain2025polaris}, trained \textit{solely} on 90~\citep{chen2025ecot-lite}. We focus on VLM steering to accommodate LIBERO's scale. 

\loose

\textbf{Real world.} 
We also aim to validate \methodName{}'s effectiveness on real robots.
We use the DROID setup to evaluate \methodName{} in the real world~\citep{khazatsky2024droid}, with $\pi_{0.5}$-DROID as our base flow VLA for steering~\citep{pi2025pi05, driess2025ki-knowledgeInsulation}. As DROID's primary challenge comes from its diversity, we choose a set of six task that require interacting with objects in scenes that admit many possible reasonable behaviors.

\subsection{\methodName{} Boosts Zero-Shot Performance on Challenging Manipulation Tasks}
\label{subsec:zero-shot-experiments}

We first test if using \methodName{} to refine VLMs' semantic guidance can boost zero-shot performance.

\textbf{Comparisons.} We compare \methodName{} with several baselines. First, we run the \textbf{base policy} without steering, to confirm that applying \methodName{} to that same policy improves performance. Second, to show that VLM actions alone are too coarse, we \textbf{directly execute the VLM's actions}, as is done in zero-shot control methods~\citep{nasiriany2024pivot, sathyamoorthy2024convoi}. Last, to show that \methodName{} is good for steering, we compare with prior policy steering methods: \textbf{partial noising}, where reference actions are interpolated with Gaussian noise before being denoised, providing biased initializations for sampling~\citep{wang2024inference-time-policy-steering, yoneda2025toNoiseAndBack} (\cref{subsec:flow-reversal}); and \textbf{sample-and-rank}~\citep{wang2024inference-time-policy-steering, nakamoto2024vgps, kwok2025robomonkey, li2025dqc-decoupledqchunking, li2026qc-reinforcementlearningactionchunking}, where the policy samples multiple actions in parallel, ranks them post-hoc with a scoring function (in our case, cosine similarity with the VLM reference action), then executes the best one. For each split, all approaches use the same flow VLA as the base policy, the same VLM system prompt, and the same Gemini-ER-1.6 VLM~\citep{grt2025geminirobotics}. See \cref{app:zero-shot-details} and \cref{app:vlm-steering-prompt}.
\loose

\begin{figure}[t]
    \centering
    \begin{minipage}{0.629\linewidth}
        \centering
        \includegraphics[width=\linewidth]{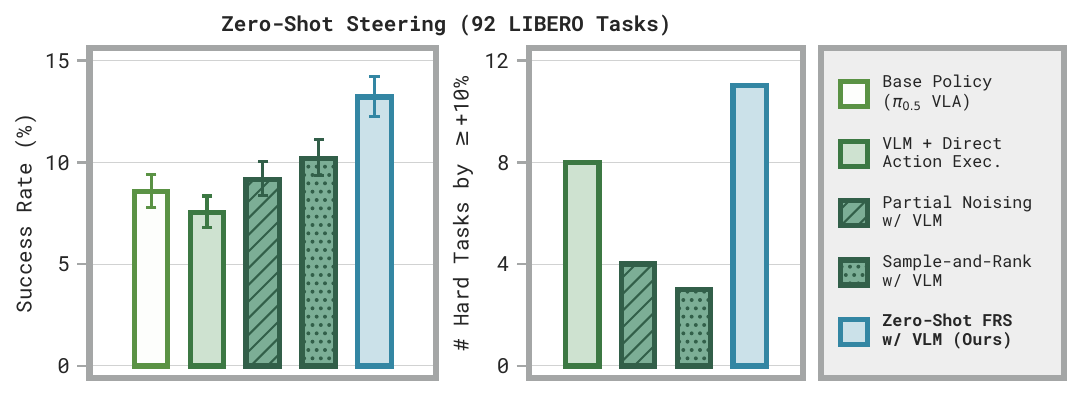}
        \caption{\textbf{Left:} Zero-shot \methodName{} improves over the base VLA across LIBERO tasks by converting coarse VLM actions into precise robot actions. \textbf{Middle:} Number of tasks improved by $\geq$10\% (where the base VLA gets $\leq$2\%). Zero-shot \methodName{} improves the most, yielding the best learning signal for RL. Directly executing VLM actions improves several, but lowers performance overall.}
        \label{fig:zero-shot-libero-results}
    \end{minipage}
    \hfill
    \begin{minipage}{0.351\linewidth}
        \centering
        \includegraphics[width=\linewidth]{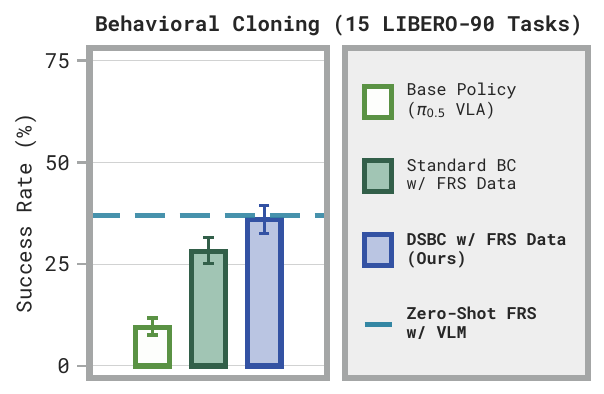}
        \caption{\methodName{} trajectories can be distilled via Diffusion Steering via Behavioral Cloning (DSBC). This matches the performance of zero-shot VLM steering, while being more effective than standard BC on the same data.}
        \label{fig:dsbc-libero-results}
    \end{minipage}
    \vspace{-1cm}
\end{figure}

\textbf{Results.} As shown in \cref{fig:zero-shot-libero-results}, \methodName{} outperforms the base policy. Critically, in 11 of the 42 LIBERO tasks where the base policy gets $\leq$ 2\% (0 or 1 success out of 50 attempts), our method yields a substantial absolute success increase of at least 10\%. While the base VLA may struggle to stumble upon even a single success, \methodName{} allows VLMs to steer the policy towards meaningful behaviors, thereby providing much earlier rewards -- and thus, beneficial learning signals -- for RL (\cref{subsec:rl-experiments}).

We also find directly executing VLM actions is ineffective. This both supports the intuition that VLMs struggle with outputting precise low-level actions zero-shot and also shows how \methodName{} is not simply reconstructing the VLM actions, but using them to steer towards better -- yet still semantically-similar -- fine-grained actions from the VLA. Finally, not only are partial noising and sample-and-rank less performant than \methodName{}, they only boost 4 and 3 hard tasks, respectively. These baselines tend to work well when the VLA already has high probability on good behaviors, not on hard tasks where success is rare, while \methodName{} is able to still learn in this case.

\subsection{\methodName{} Enables Diffusion Steering via Behavior Cloning}
\label{subsec:dsbc-experiments}

We now show how good trajectories from \methodName{} yield expert noise actions, which can be distilled via DSBC. We focus on online DSBC here, and present offline DSBC LIBERO results in \cref{app:offline-dsbc-libero}.

\textbf{Comparisons.} We start by considering online DSBC on zero-shot \methodName{} data.
Alongside the \textbf{base policy} and \textbf{zero-shot VLM \methodName{}} as baselines, we also compare against running \textbf{standard BC} on the \methodName{} successful trajectories (using the same small architecture as the DSBC noise policy). We note that \methodName{}'s successful rollouts can be distilled back into the full VLA as well, but doing so uses much more compute than training an auxiliary policy. See \cref{app:dsbc-details}.

\begin{figure}[t]
    \centering
    \includegraphics[width=1\linewidth]{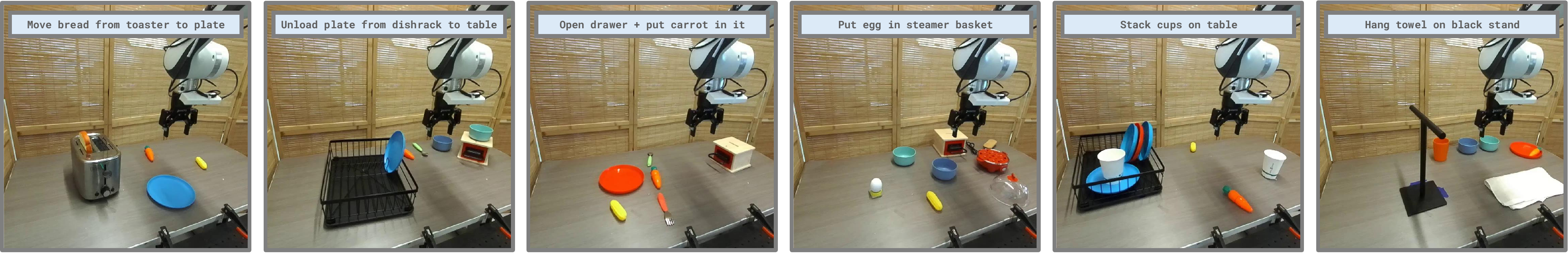}
    \includegraphics[width=1\linewidth]{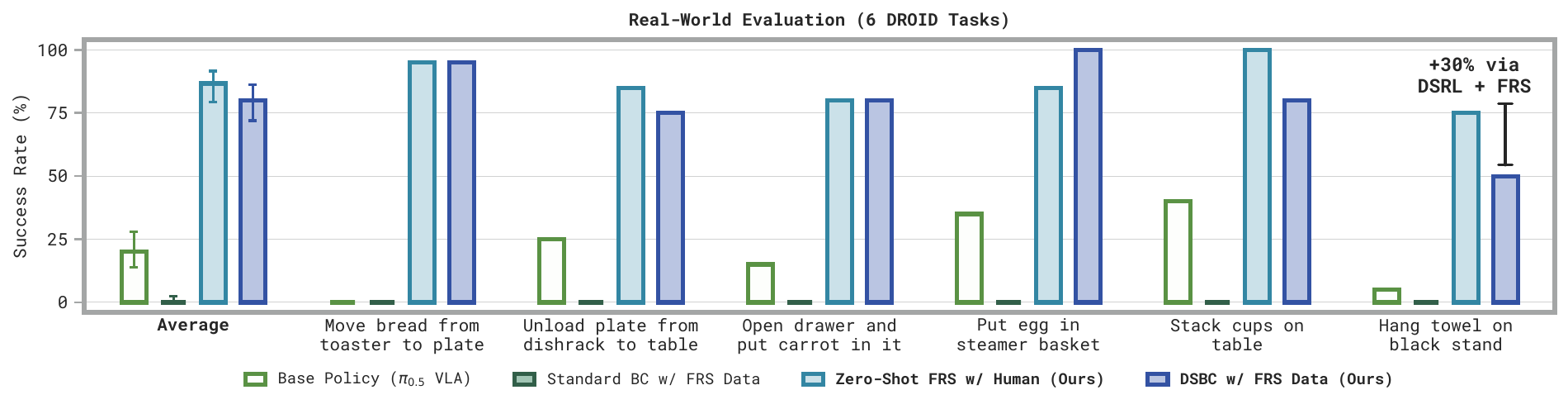}
    \vspace{-0.7cm}
    \caption{
    DSBC boosts performance on real-world tasks when trained with just 10 \methodName{} rollouts, while standard BC \textit{completely fails} in this data regime. DSBC rollouts can also bootstrap RL, as in the towel hanging task.
    }
    \label{fig:real-results}
    \vspace{-0.5cm}
\end{figure}

\textbf{Results.} As shown in \cref{fig:dsbc-libero-results}, DSBC distills the zero-shot gains of \methodName{}, improving over the base VLA. One empirical benefit of DSBC is that when the noise policy makes mistakes and enters out-of-distribution states, it is often able to recover. We posit that, while the noise policy's actions may be bad at these OOD states, the VLA treats those noises akin to its noise prior, mapping them to ``reasonable'' in-distribution actions. Essentially, \textit{the DSBC noise policy is implicitly robust against compounding error, as it ``falls back'' to the VLA's behavioral prior in unfamiliar states}. As DSBC can rely on the VLA's action prior, it is better to run BC on noise actions than regular actions. 
In LIBERO, DSBC outperforms standard BC (though, as LIBERO has little randomization, memorizing a small dataset can still work~\citep{zhou2025liberoPro, wang2026liberox}).
This difference is more salient in the real world, where \textit{standard BC with a small flow policy completely fails} (\cref{fig:real-results}).

DSBC is also sample-, compute-, and time-efficient. In LIBERO, it trains on only 18 rollouts per task on average. On real robots, it needs just 10 rollouts per task to achieve high performance (\cref{subsec:real-experiments}). The policy is likewise small -- in total, training takes around 1 GB of GPU memory (as the VLA does not need to be loaded during training), whereas fine-tuning a full VLA requires hundreds of GBs. Finally, due to the model and data size, DSBC policies take under a minute to train.\loose

\subsection{\methodName{} Accelerates and Improves Reinforcement Learning}
\label{subsec:rl-experiments}

\begin{figure}[t]
    \centering
    \includegraphics[width=1\linewidth]{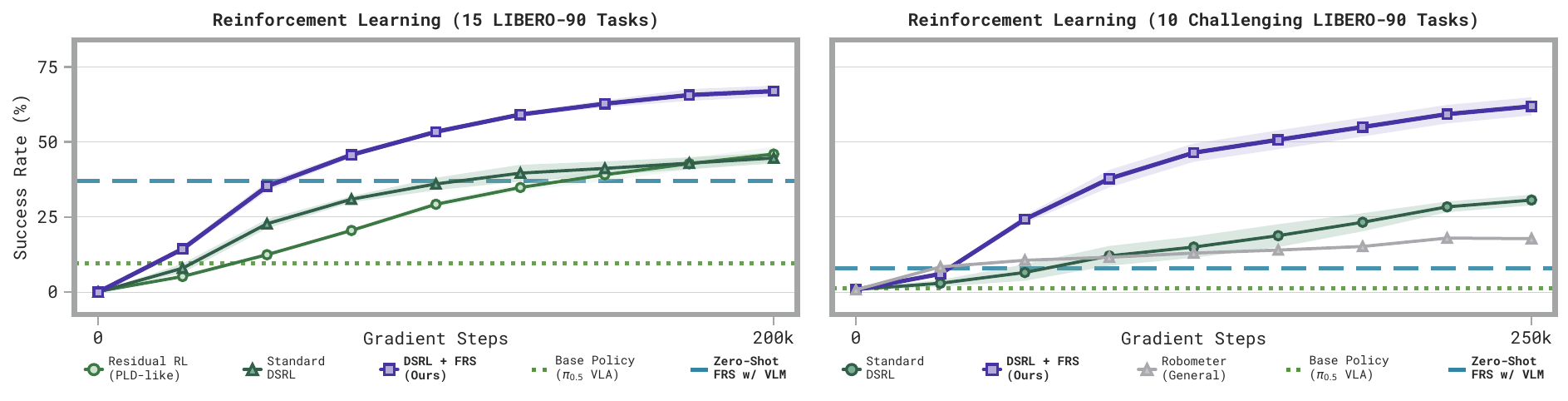}
    \vspace{-0.6cm}
    \caption{
    \textbf{Left:} DSRL + \methodName{} improves upon standard generalist RL methods~\citep{xiao2025pld-probeLearnDistill, wagenmaker2025dsrl}. \textbf{Right:} Even if \methodName{} struggles, warmstarting with even one  \methodName{} success improves RL on tasks where the base policy gets near-zero.}
    \label{fig:rl-libero-results}
    \vspace{-0.9cm}
\end{figure}

We finally aim to show how \methodName{} can be used with RL to learn from experience. This allows it to surpass both the fixed performance of zero-shot \methodName{} and distilling \methodName{}'s data with DSBC.

\textbf{Settings and comparisons.} We run RL in two LIBERO-90 settings. First, we consider the 15 tasks from \cref{subsec:dsbc-experiments} where zero-shot VLM \methodName{} is especially effective (yielding $\geq$10\% improvement), and run DSRL + \methodName{} by selecting 20 random \methodName{} rollouts to prefill the replay buffer (in place of some initial prefill rollouts). Our baselines are thus two standard VLA RL methods that do \textit{not} use \methodName{} data: (1) \textbf{standard DSRL}~\citep{wagenmaker2025dsrl} and (2) \textbf{residual RL} (akin to PLD ~\citep{xiao2025pld-probeLearnDistill}). Second, we consider 10 harder LIBERO-90 tasks where the base VLA nearly always fails \textit{and} zero-shot VLM \methodName{} achieves only 8\%. This tests if \methodName{} is useful for RL, \textit{even if steering rarely succeeds}. 
We thus run DSRL + \methodName{} with only one successful steered trajectory, which can take upwards of 50 trials, given the tasks' difficulty.
As densifying rewards is another way to guide RL with VLMs, we also compare against using \textbf{RoboMeter} as a reward model. 
All methods run RL on a small policy to steer a VLA, albeit in different ways. We thus control for the VLA, the small policy's architecture, and the underlying RL algorithm (SAC). See \cref{app:dsrl-frs-details}.\loose

\textbf{Results.} DSRL + \methodName{} is the most effective and sample-efficient RL method in both our LIBERO settings. As shown in \cref{fig:rl-libero-results} (left), running RL with \methodName{} rollouts as prior data yields significant gains over standard RL, enabling both faster learning and higher final success rate across 15 tasks. For our second setting, where the base VLA has success rate near 0\%, DSRL + \methodName{} again enables effective improvement as \cref{fig:rl-libero-results} (right) shows. Naive DSRL struggles to learn -- only reaching a final success rate of around 30\%, likely due to the poor performance of the base policy. By leveraging VLM \methodName{} to direct the learner to successful behaviors in early stages of learning, DSRL + \methodName{} is able to overcome this, quickly improving and converging to a significantly higher final success rate.

\subsection{\methodName{} is Practical and Effective for Real-World Manipulation}
\label{subsec:real-experiments}

Lastly, we show \methodName{} is effective in real-world generalist manipulation (\cref{fig:real-results}). As steering offloads the effort of producing fine actions to the policy, \methodName{} lets humans solve tasks while only giving very crude feedback (i.e., one Cartesian directional action per action chunk), compared to the dense supervision provided during teleoperation. The successful human-steered trajectories can then be used for DSBC. Across six tasks that the base VLA struggles on, we find that corresponding DSBC policies boost average absolute performance by 60\% by training on just 10 successful human \methodName{} rollouts per task. Equivalent \textit{standard} BC flow policies trained on the \methodName{} robot actions completely fail to learn these tasks in this data regime, as they cannot inherently rely on the VLA action prior in unfamiliar situations (see \cref{subsec:dsbc-experiments}).
As in the LIBERO DSBC experiments, each training run takes under a minute and requires around 1 GB of GPU memory. Lastly, we show how DSBC can be used to bootstrap trajectories for simple online RL, as we demonstrate on the challenging towel-hanging task by boosting performance from 5\% base, to 50\% via DSBC, and then to 80\% post-RL (\cref{app:real-world}).

\begin{wrapfigure}[8]{r}{0.5\linewidth}
  \centering
  \vspace{-1.5em}
  \begin{minipage}{0.49\linewidth}
    \centering
    \includegraphics[height=2.5cm]{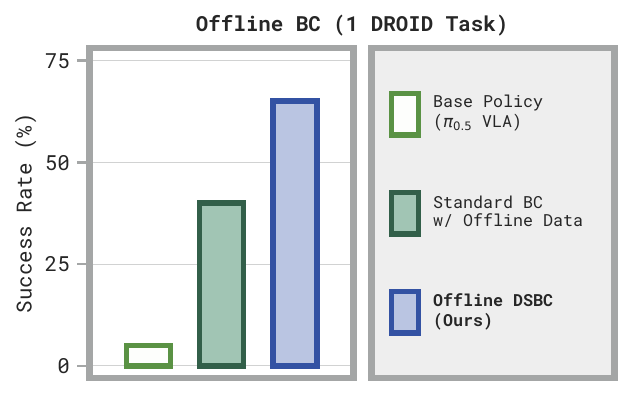}
  \end{minipage}
  \hfill
  \begin{minipage}{0.49\linewidth}
    \centering
    \includegraphics[height=2.375cm]{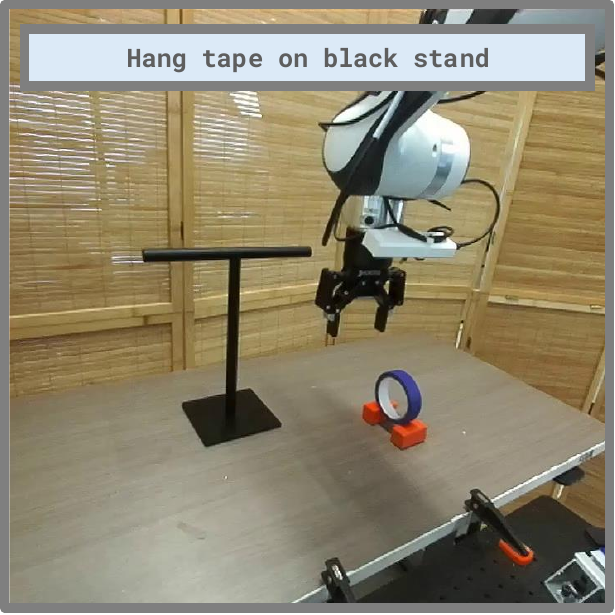}
  \end{minipage}
  \vspace{-0.2cm}
  \caption{Offline DSBC enables noise policy learning from \textit{standard} robot data.}
  \label{fig:offline-dsbc}
\end{wrapfigure}
\textbf{Real-world offline DSBC.} Offline DSBC can use \textit{standard} robotic trajectories (i.e., with only robot actions saved, and no noises),
enabling learning with more optimal demonstrations than what is possible through a coarse steering interface. 
We test this with a real-world task that na\"{i}ve directional steering struggles on due to imprecision. We collect 20 episodes of the task ``hang the tape on the stand'' via regular teleoperation (i.e., without noises). Then, we use $\pi_{0.5}$-DROID flow reversal to augment all episodes' actions with their corresponding noises, which DSBC learns from. This noise outperforms the base VLA and standard BC, which struggles to learn precise, temporally-coherent behaviors in our low-data regime (\cref{fig:offline-dsbc}). This validates flow reversal as a simple way for noise policies to make use of standard offline robot data \textit{without} noises.

%% file: sections/06-discussion.tex
\section{Discussion}
\label{sec:discussion}

We introduce Flow Reversal Steering (\methodName{}), a way to convert coarse semantic guidance into precise actions by reversing flow generalist policies. Through extensive simulated and real-world tasks with state-of-the-art VLAs, we show how \methodName{} allows reasoners, like humans and VLMs, to guide policies towards reasonable behaviors for novel tasks. This enables rapid policy learning through our novel Diffusion Steering via Behavior Cloning (DSBC) method or by bootstrapping DSRL. While we showed \methodName{}'s effectiveness, some parts are limited. 
We hope that \methodName{} provides an alternative paradigm for efficiently improving generalist policies, where learning is not only guided by optimizing reward functions, but task-relevant semantic knowledge as well.

%% file: sections/appendices.tex
\clearpage
\appendix

\section{Additional Related Work}
\label[appendix]{app:related_additional}

\paragraph{Flow and diffusion inversion in other domains.}
While this work has focused on applying flow reversal to steering robotic control policies, flow and diffusion reversal (or ``inversion'') approaches have found applications in a variety of other domains. In particular, in image domains, inversion-based approaches have enabled controllable image editing both for image generation diffusion models sampled with deterministic DDIM \cite{song2020denoising} sampling \cite{mokady2023null, kim2022diffusionclip, tumanyan2023plug, wallace2023edict, hertz2022prompt, su2022dual} and flow models \cite{rout2024semantic, wang2024taming, deng2024fireflow, avrahami2025stable,jiao2025uniedit}.
Other applications of flow and diffusion inversion in image domains include image restoration \cite{chihaoui2024blind}, watermarking \cite{yang2024gaussian}, and detection of AI-generated images \cite{wang2023did, cazenavette2024fakeinversion}.
Outside of image domains, diffusion inversion has also been applied to editing audio signals \cite{manor2024zero, liu2024medic}. While the techniques utilized in these works are similar to our flow inversion approach, the applications differ substantially---our work shows that flow inversion can lead to significant performance improvements in robotic domains.

\section{Illustrative Examples of Flow Reversal Steering}
\label[appendix]{app:illustrative-flow-reversal-expts}

\subsection{Idealized FRS Example}

As the distribution of large-scale flow matching policies such as those used in the paper precludes calculation of log-probabilities of the pre- and post-FRS actions, we consider an idealized setting where the action distribution is fixed to a mixture of Gaussians to allow for exact calculation of the dynamics of FRS. We are interested in three questions:

\begin{enumerate}
    \item What does the change in coarse action distribution look like qualitiatively?
    \item Does FRS make actions more in-distribution? We expect a positive \textbf{log-density ratio}, defined as (total log-likelihood post-FRS) - (total log-likelihood pre-FRS). 
    \item Does FRS keep actions near their respective modes? We should expect FRS to have an average movement of all coarse steering actions by a small amount in action space.
\end{enumerate}

\begin{figure}[t]
    \centering
    \includegraphics[width=0.9\linewidth]{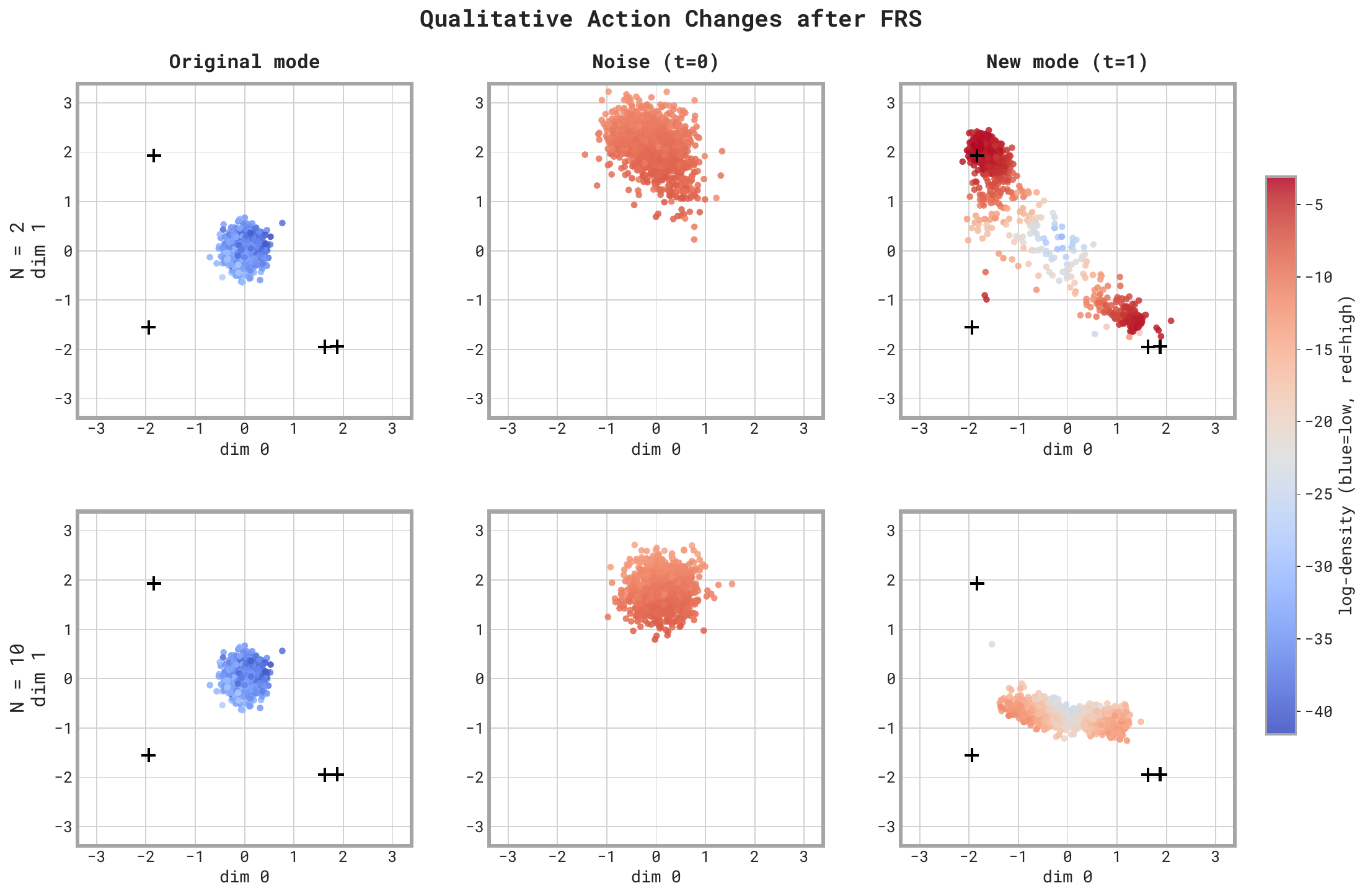}
    \caption{
    In a simplified setting, FRS brings actions from the original out-of-distribution mode into more in-distribution actions as measured by log-likelihood. Fewer steps in the Euler approximation results in actions that are more in-distribution but farther from the original mode.
    }
    \label{fig:one-step-results}
\end{figure}

To empirically analyze these questions, we design a setting with four ``action modes'' in 7-dimensional space, all with standard deviation 0.5 with means in $[-2, 2]^7$. Pairwise distances between modes are fixed to be at least $4$, and the norm of each center is also at least $4$. This ensures the modes are separate and the area around the origin is out-of-distribution. We train a flow-matching model with 10,000 sampled points from this true action distribution using a three-layer MLP. 

As a proxy for a steered mode, we sample 1,000 actions from $N(0, 0.2)$, and pass it through $K=[2, 3, 5, 10, 20, 50, 100]$ steps in FRS. We expect that FRS with low $K$ will have higher average movement of each action, but also larger log-probability ratio.

\begin{figure}[t]
    \centering
    
    \begin{subfigure}[t]{0.49\linewidth}
        \centering
        \includegraphics[width=\linewidth]{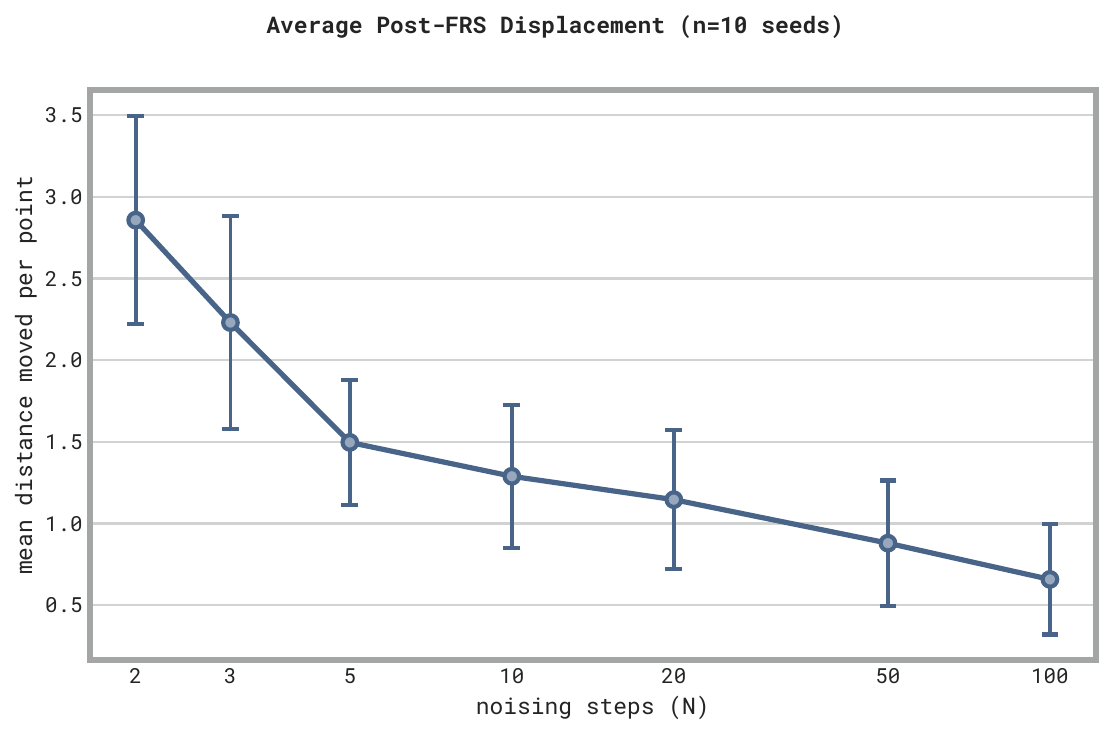}
        \label{fig:distance-moved}
    \end{subfigure}
    \hfill
    \begin{subfigure}[t]{0.49\linewidth}
        \centering
        \includegraphics[width=\linewidth]{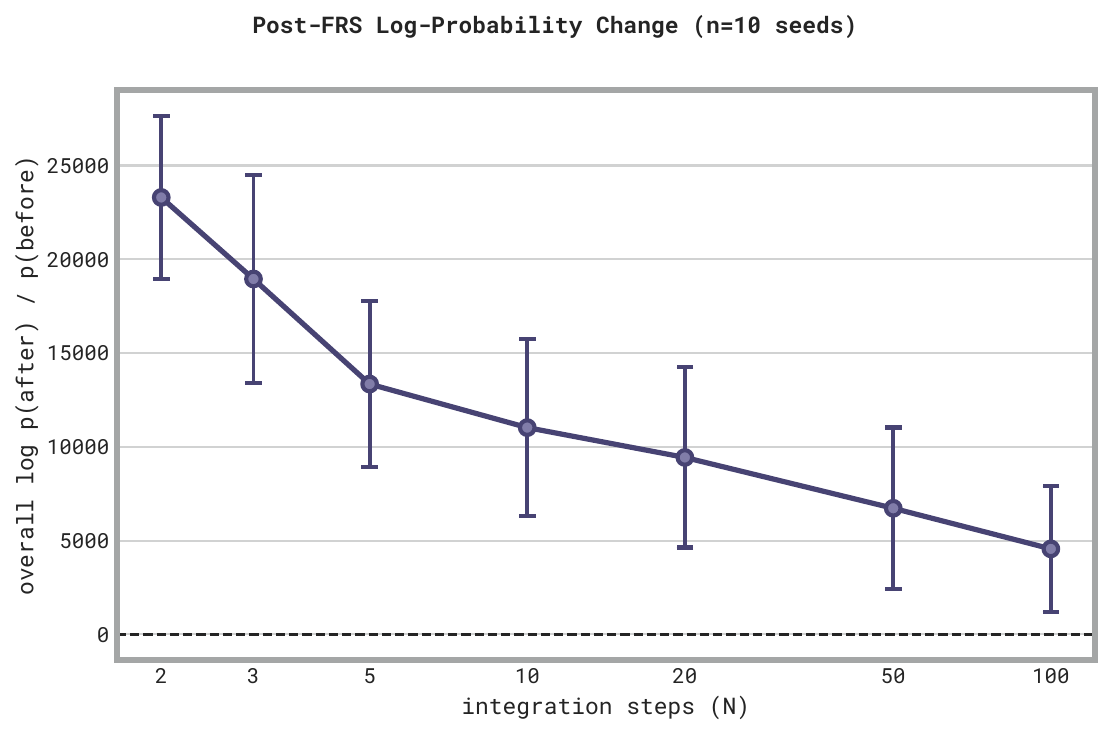}
        \label{fig:logprob-ratio}
    \end{subfigure}

    \caption{
    Fewer steps during FRS result in higher log-density ratio (indicating actions that are more in-distribution on average) but also higher displacement relative to the number of noising steps. Note that on average in this simplified setting, any point from our steering ``coarse action'' should be expected to have distance around $4$ from any mode. We choose $10$ noising steps in our robotics experiments as a good balance between action fidelity and improved log-density ratio.
    }
    \label{fig:distance-results}
\end{figure}

Qualitatively, we visualize $K=2, 10$ results in \cref{fig:one-step-results}. We see that although the original mode has very low-density actions, after noise inversion both the noise seeds and the transformed actions have much higher log-probabilities compared to the original mode. For $K=2$, we move further towards high density compared to $K=10$. We conjecture this occurs because integration error during the learned flow process tends to push the policy towards areas with higher probability density, with fewer steps leading to more integration error.

Quantitatively, we confirm over 10 seeds in \cref{fig:distance-results} that increasing the number of noising steps results in a decrease in the mean distance moved per point, but also a decrease in round-trip log-density ratio, meaning actions are staying closer to the mode but also ending up less in-distribution. With infinite steps, we expect the overall ratio to converge to zero and the displacement to also converge to near-zero. We choose $10$ noise steps of size $0.1$ in our simulation and real-world robotic experiments to balance these two concerns. We leave optimization of this technique to future work.

\begin{figure}[t]
    \centering

    \begin{subfigure}[t]{\linewidth}
        \centering
        \includegraphics[width=\linewidth]{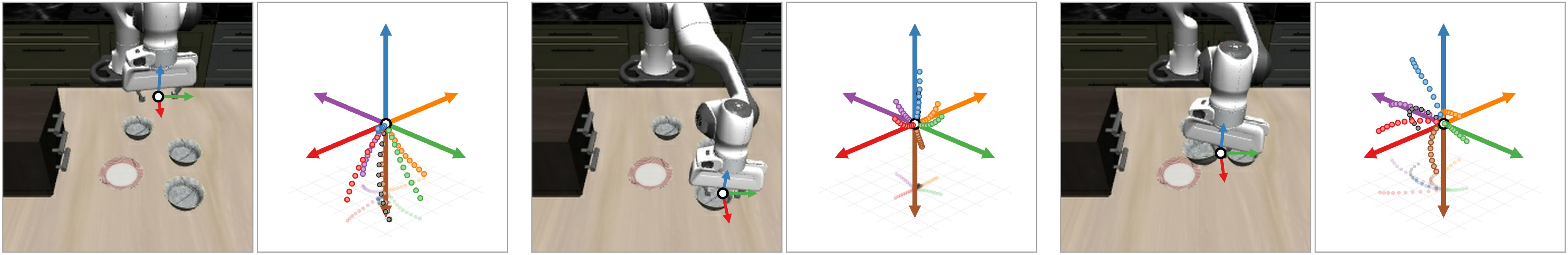}
        \caption{The VLA initially can be steered toward reaching any of the 3 bowls. Once one is grasped, the steered actions are biased up to lift the bowl. Finally, steering can either push the policy to put the bowl on the plate (correct) \textit{or} atop the cabinet (blue, incorrect) -- both are ``reasonable'' behaviors.}
        \label{fig:libero-steering-examples1}
    \end{subfigure}
    \begin{subfigure}[t]{\linewidth}
        \centering
        \includegraphics[width=\linewidth]{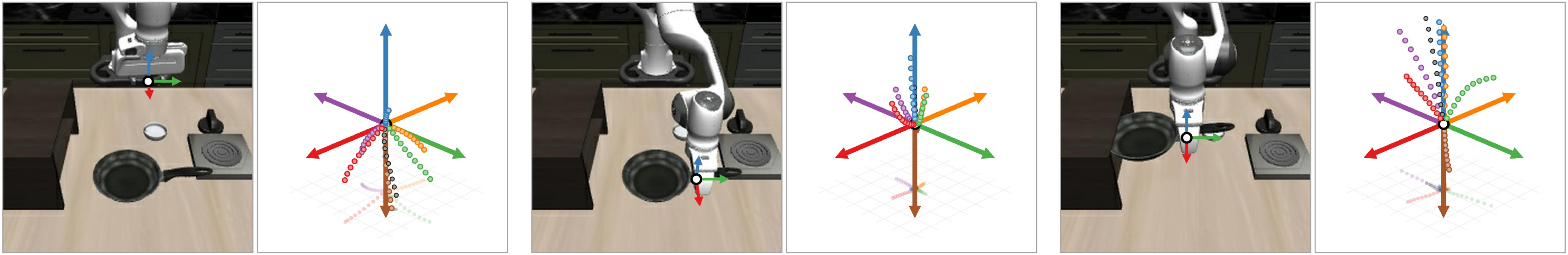}
        \caption{Initial steering is flexible, as reaching for the bowl, knob, or pan are all reasonable. Grasps causes bias towards lifting, and once lifted, main action modes are towards the shelf, above the shelf, or to the stove (green).}
        \label{fig:libero-steering-examples2}
    \end{subfigure}
    \begin{subfigure}[t]{\linewidth}
        \centering
        \includegraphics[width=\linewidth]{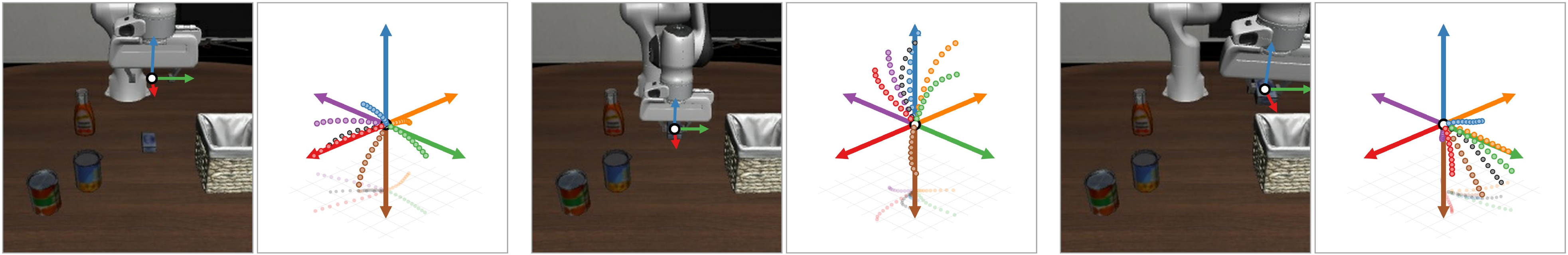}
        \caption{Initial steering is again very flexible, as moving to any object left of the basket is reasonable. Grasping biases samples upwards. Finally, once close to the basket, samples are biased towards moving to place the object inside.}
        \label{fig:libero-steering-examples3}
    \end{subfigure}
    \begin{subfigure}[t]{\linewidth}
        \centering
        \includegraphics[width=\linewidth]{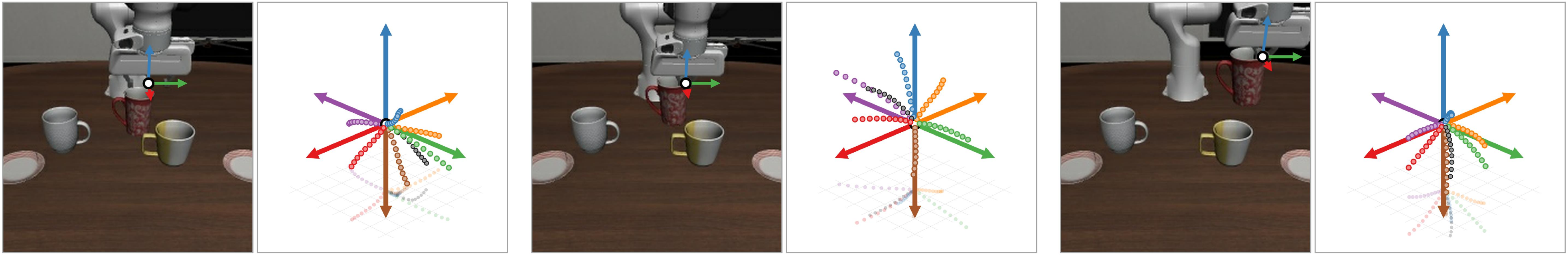}
        \caption{Initial steering can move towards any of the three mugs (but all are biased to moving down). Once grasped, there is significant variation, as placing on either plate is reasonable (left and right with purple and green have strong reconstructions). Once above a plate, actions are biased toward moving down for placing.}
        \label{fig:libero-steering-examples4}
    \end{subfigure}

    \caption{Examples of running \methodName{} in LIBERO. The colored arrows are the cardinal reference actions, and the dotted points are the \methodName{} actions they result in (black is the policy's base action). We display these actions in 3D side views for clarity, but note that the red, green, and blue axes correspond to the ones overlaid on the robot images (forward, right, and up respectively).
    In general, the actions move in the rough direction specified by the reference, but are more akin to actions drawn from the VLA's distribution and are highly dependent on the current observation.}
    \label{fig:libero-steering-examples}
\end{figure}

\subsection{Robot FRS Example}
In \cref{fig:libero-steering-examples}, we show actual examples of running \methodName{} in LIBERO-90. The base policy is $\pi_{0.5}$-LIBERO, which we reiterate was not trained on 90 -- it thus just places a wide distribution over ``reasonable'' actions. We steer it with directional actions which we display as solid arrows, both overlaid on the robot observation \textit{and} in auxiliary 3D plots (viewed at a three-quarters angle). Naturally, these are very coarse reference actions (executing them directly on the robot is unlikely to be effective).
The corresponding colored dotted lines are the actions produced by running \methodName{} on the coarse arrow actions, while the black represents a sample from the base VLA (which is often incorrect for the task, as these settings are OOD). We always use 10 integration steps ($h=0.1$).

As desired, the steered actions roughly follow the directional reference actions. However, they do not perfectly reconstruct them -- instead resembling samples from a nearby action mode of the base generalist VLA. They also depend heavily on the observation: e.g., starting frames tend to bias steering downward (as the policy usually moves down to grasp), or when grasping an object the actions are biased upwards to lift it. Additionally, when different steering directions are close to different ``reasonable'' behaviors (grasping different objects, placing a held object in different reasonable places), then those directions map to actions that correspond to those behavioral modes.

\section{What is Going On in Flow Policy Noise Space?}
\label[appendix]{app:noise-space-details}

\begin{figure}[t]
    \centering
    \includegraphics[width=\linewidth]{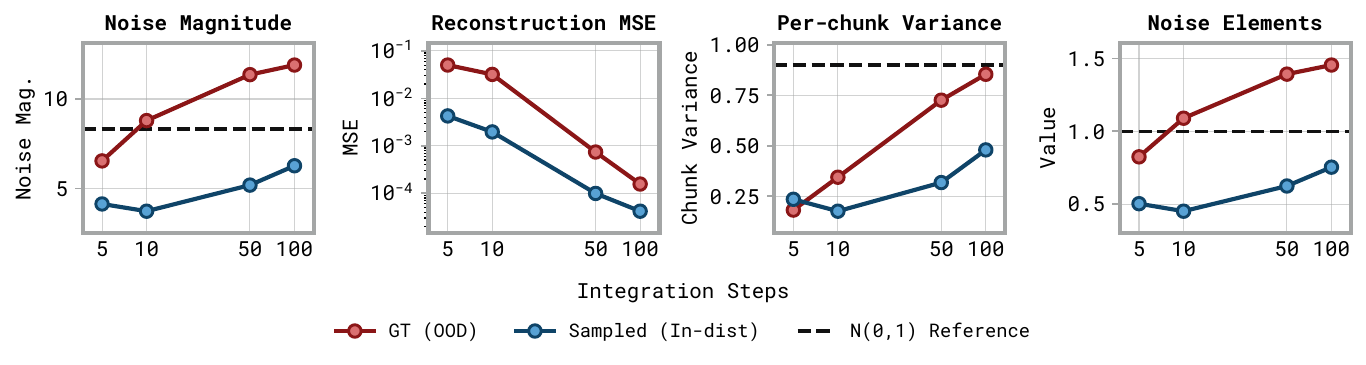}
    \vspace{-0.75cm}
    \caption{How mean noise magnitude, reconstruction MSE, noise value distribution, and per-chunk variance change with different numbers of integrations steps when applying \methodName{} to the LIBERO-90 dataset.}
    \label{fig:libero90-per-step-results}
    \vspace{-0.5cm}
\end{figure}

\begin{figure}[t]
    \centering

    \begin{subfigure}[t]{\linewidth}
        \centering
        \includegraphics[width=\linewidth]{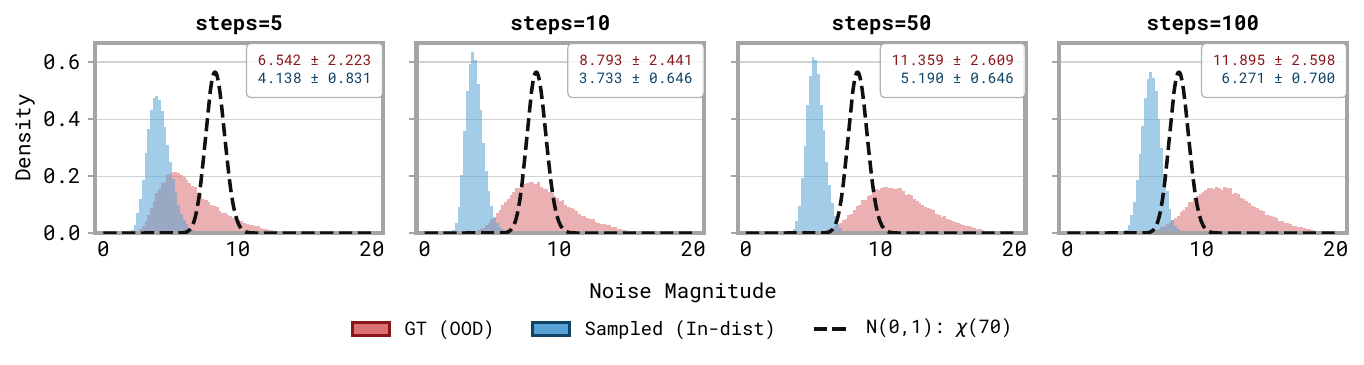}
        \vspace{-0.75cm}
        \caption{Magnitude of noises produced by running \methodName{} on LIBERO-90 data. Black is the distribution if noises were drawn from $\mathcal{N}(0, I)$, following a $\chi$ distribution.}
        \label{fig:libero90-noise-mag}
    \end{subfigure}
    
    \begin{subfigure}[t]{\linewidth}
        \centering
        \includegraphics[width=\linewidth]{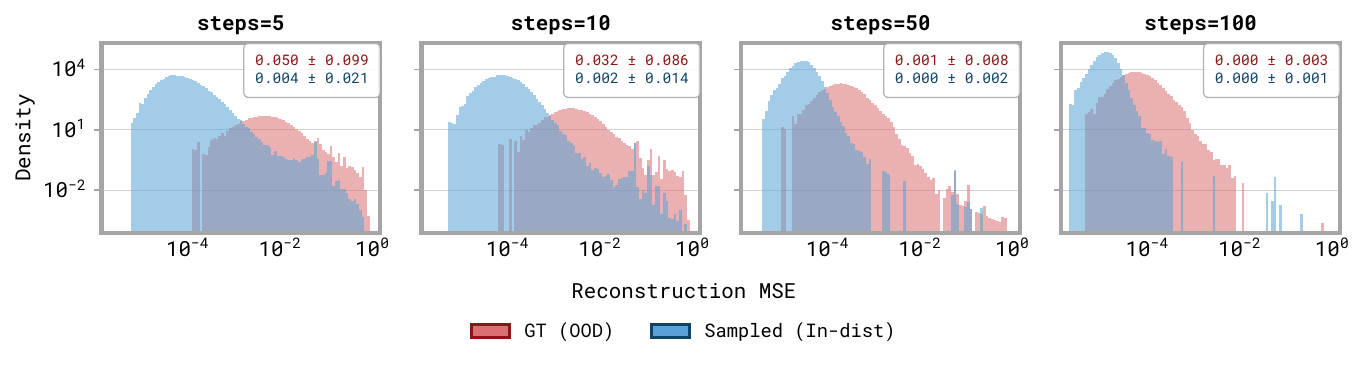}
        \vspace{-0.75cm}
        \caption{Reconstruction MSE when running \methodName{} on LIBERO-90 data.}
        \label{fig:libero90-recon-mse}
    \end{subfigure}
    
    \begin{subfigure}[t]{\linewidth}
        \centering
        \includegraphics[width=\linewidth]{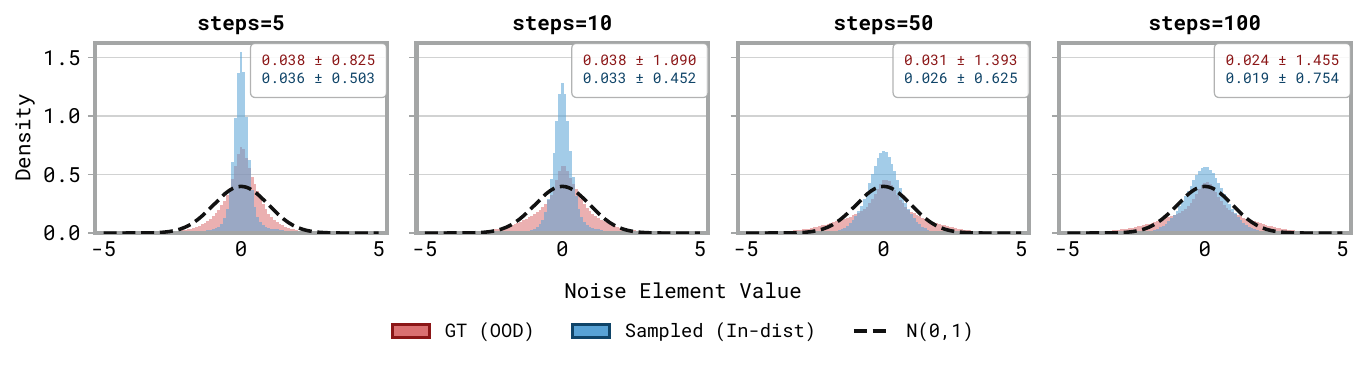}
        \vspace{-0.75cm}
        \caption{Distribution of noise values when running \methodName{} on LIBERO-90 data. Black is the distribution if noises were drawn from $\mathcal{N}(0, I)$.}
        \label{fig:libero90-element-values}
    \end{subfigure}

    \begin{subfigure}[t]{\linewidth}
        \centering
        \includegraphics[width=\linewidth]{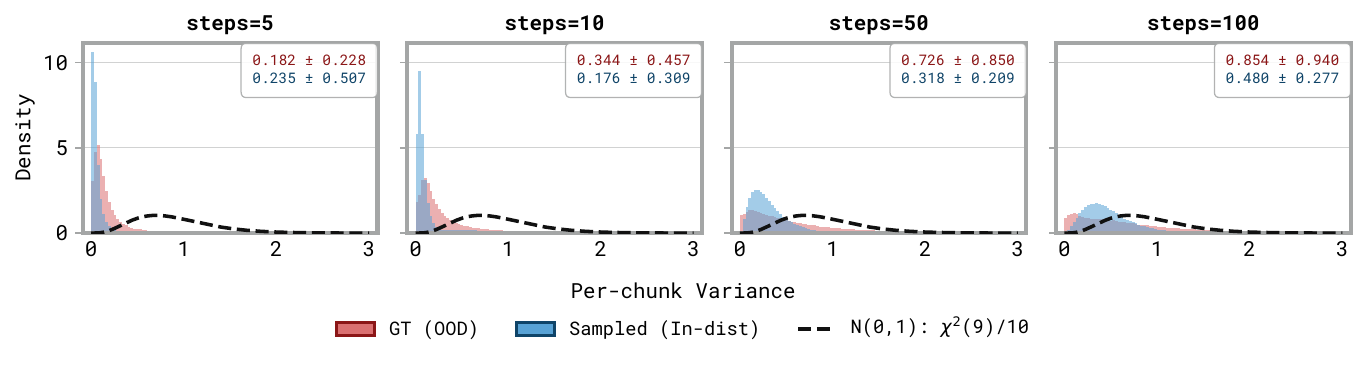}
        \vspace{-0.75cm}
        \caption{Variance of noise values across the chunk axis when running \methodName{} on LIBERO-90 data.}
        \label{fig:libero90-chunk-var}
    \end{subfigure}
    \caption{Running \methodName{} on the full LIBERO-90 dataset with varying integration steps. Black is if noises were drawn from $\mathcal{N}(0, I)$, empirical mean and standard deviation are also reported.}
    \label{fig:libero90-noise-analysis}
\end{figure}

\begin{figure}[t]
    \centering
    \includegraphics[width=\linewidth]{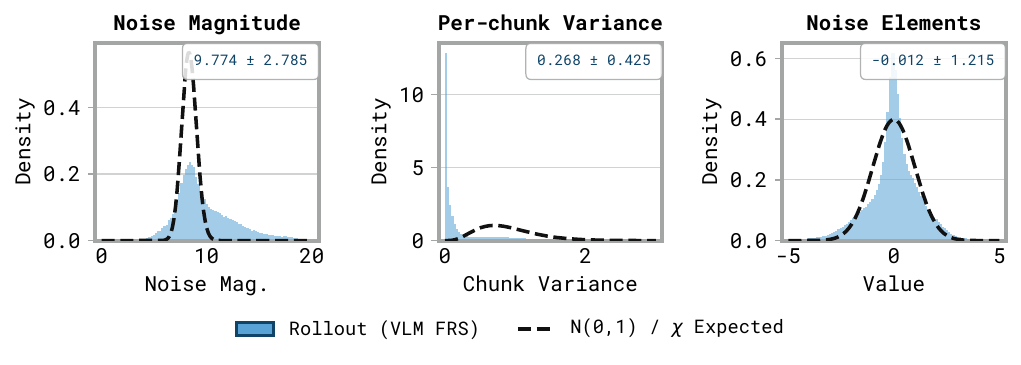}
    \caption{Analysis of noise statistics from our zero-shot VLM \methodName{} experiments on LIBERO-90.}
    \label{fig:rollout-noise-results}
\end{figure}

In testing \methodName{}, we obtained several insights into the structure of flow policies' noise space. We ran several ablative and auxiliary experiments that allow us to form a more complete characterization of noise space. For convenience, we first summarize all results in \cref{app:summary-noise-space-insights}. These experiments also motivate certain design choices when running \methodName{} in practice, which we outline in \cref{app:noise-space-takeaways}. 

\subsection{Summary of Noise Space Insights}
\label[appendix]{app:summary-noise-space-insights}

\cref{app:ood-actions}:
\begin{itemize}
    \item When running flow reversal, more likely noises (i.e., ones closer to 0) tend to map to more in-distribution actions.
    \item Flow reversal is better able to find noises that reconstruct to more in-distribution actions (lower reconstruction MSE).
    \item Increasing number of integration steps for flow reversal leads to more accurate reconstruction (due to lower integration error), but at the cost of going more OOD in noise space.
\end{itemize}

\cref{app:expert-noise}:
\begin{itemize}
    \item \methodName{} finds good regions of noise space. Noises \textit{near} good noises also map to good actions.
    \item Noises corresponding to padding elements in action chunks do not seem to affect denoising much.
\end{itemize}

\cref{app:noise-averaging}:
\begin{itemize}
    \item When applying flow reversal, action chunks tend to map to temporally correlated noise chunks, with lower variance across the chunk axis than if sampled from $\mathcal{N}(0, I)$.
    \item When an \textit{action} is averaged and repeated over the chunk axis, it can still find good noises via flow reversal (at least for delta end effector control, where averaging is sensible).
    \item Likewise, when a \textit{noise} produced by flow reversal is averaged and repeated over the chunk axis, it can still denoise to good actions.
\end{itemize}

\subsection{\methodName{} on In- and Out-of-Distribution Actions}
\label[appendix]{app:ood-actions}
We first investigate how in- vs. out-of-distribution actions affect flow reversal.
We use the LIBERO-90 filtered dataset released by \citet{chen2025ecot-lite} in conjunction with the $\pi_{0.5}$-LIBERO checkpoint released by \citet{pi2025pi05}. 
For every 10th frame in that dataset, we procure two action chunks: (1) the ground-truth actions from the dataset (As $\pi_{0.5}$-LIBERO was \textit{not} trained on 90, the actions from the LIBERO-90 dataset are OOD for it) and (2) a single sampled action from $\pi_{0.5}$-LIBERO, given each frame (which is naturally ID).

Now that we have ID and OOD actions, we run flow reversal for both sets with different numbers of integration steps. Specifically, we use $\{ 5, 10, 50, 100 \}$ integration steps. Both noising with flow reversal and denoising with standard flow matching use the same number of steps in each of the four conditions.
See \cref{fig:libero90-per-step-results} for aggregated trends and \cref{fig:libero90-noise-analysis} for more details.

\textbf{OOD actions map to less likely noises}. Intuitively, we find that ID actions consistently map to noise vectors of smaller $L_2$ magnitude (\cref{fig:libero90-noise-mag}). Since noises are typically sampled from $\mathcal{N}(0, I)$, smaller magnitude noises are more likely to be sampled from the noise prior. Flow reversal's noises thus seem to act as a pseudo proxy for how in-distribution the reference actions are. The fact that the noises from zero-shot VLM \methodName{} also tends to have higher noise magnitudes supports this as well, as the straight directional actions emitted by the VLM are OOD (\cref{fig:rollout-noise-results}).

When we denoise the outputs of flow reversal, we unsurprisingly find that OOD actions have higher reconstruction MSE than the ID ones (\cref{fig:libero90-recon-mse}). It is thus \textit{easier} to find noises that reconstruct to ID actions than OOD ones. 

\textbf{Integration steps control noise likelihood and reconstruction error.}
For both ID and OOD actions, we find that fewer integration steps results in lower noise magnitudes (\cref{fig:libero90-noise-mag}), but higher reconstruction MSE (\cref{fig:libero90-recon-mse}). As the number of steps increases, reconstruction MSE decreases (the reconstruction is \textit{better}), but the flow reversal noises grow in magnitude (and, indeed the overall variance of noise elements~\cref{fig:libero90-element-values}. See \cref{fig:libero90-per-step-results} for aggregated trends.

The reconstruction trend is expected: more integration steps means there is less integration error, so action reconstructions should get better. 
In turn, we interpret the noise magnitude trend as suggesting that, by using more integration steps, flow reversal finds more OOD noises that more closely denoise to the reference action.

We thus opt to use 10 integration steps for all our main experiments -- not only is this the default for OpenPi flow-matching policies~\citep{black2024pi0, pi2025pi05}, but using more could result in flow reversal finding more OOD/high-magnitude noises, which may be harder to use when learning noise policies for DSBC or DSRL + \methodName{}.

\subsection{Better Reference Actions Enable Better Steering Performance}
\label[appendix]{app:expert-noise}

\begin{figure}[t]
    \centering
    \includegraphics[width=\linewidth]{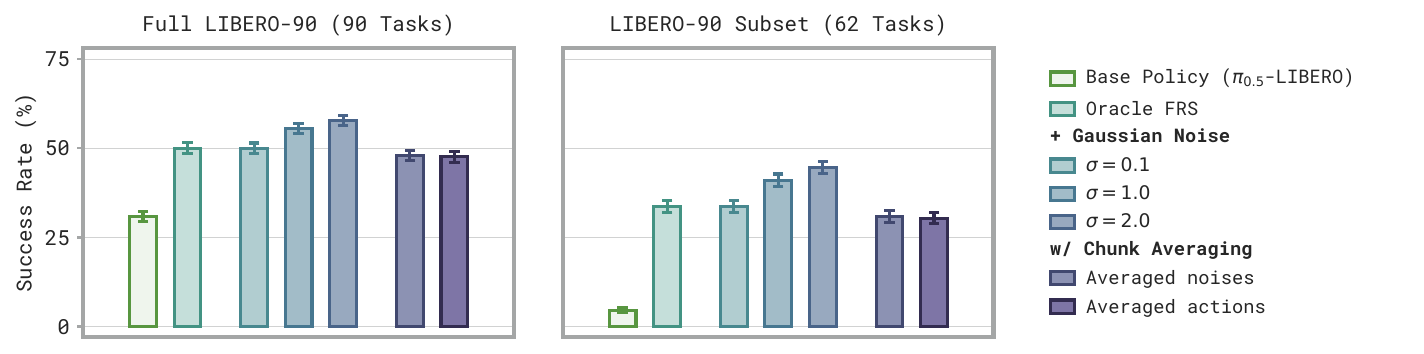}
    \caption{LIBERO-90 success rates from running \methodName{} using an \textit{oracle} VLA to steer, as well as ablations. The 62-task subset is the one used for zero-shot VLM \methodName{} experiments.}
    \label{fig:oracle-results}
\end{figure}

Perhaps unsurprisingly, \methodName{}'s zero-shot performance improves as the quality of the steering reference actions improves -- e.g., if we replace our coarse cardinal steering actions with high-quality, fine-grained action chunks from an expert policy.

\textbf{Steering with fine-grained oracle actions.} We again consider $\pi_{0.5}$-LIBERO achieves 30.8\% on all of LIBERO-90 and 4.5\% on the challenging task subset we consider in \cref{subsec:zero-shot-experiments}.
Rather than using coarse actions from a VLM for steering it, we instead use the expert $\pi_{0.5}$ policy trained by \citet{jain2025polaris}, which was trained only on LIBERO-90 and achieves 93.9\% performance on that benchmark (92.8\% on the challenging tasks). The latter VLA thus acts as as ``oracle'' for LIBERO-90. See \cref{fig:oracle-results} for all results and ablations.

We find that using \methodName{} with expert VLA steering achieves 50\% on all of LIBERO-90 (33.7\% on the challenging split, higher than the 11.1\% from VLM steering). This does not fully recover the oracle's performance, though still shows that using better reference actions increases zero-shot \methodName{} performance.

\textbf{\methodName{} finds good \textit{regions} of noise space.} We also try adding 0-mean Gaussian noise to the noise deterministically produced by \methodName{}. For reference action $a_1$ sampled from the oracle, we run \methodName{} $\hat{a}_0 \leftarrow \mu^{-1}_\theta(a_1, o)$ as normal, then execute action $\hat{a}_1$ produced via:
\begin{equation}
     \hat{a}_1 \leftarrow \mu_\theta(\hat{a}_0 \textcolor{red}{+ \sigma \epsilon}, o); \ \textcolor{red}{\epsilon \sim \mathcal{N}(0, I)}
\end{equation}
We test $\sigma \in \{ 0.1, 1, 2 \}$. As shown in \cref{fig:oracle-results}, while 0.1 keeps performance the same, we unexpectedly found that using noise scales of 1 or 2 actually substantially boost performance. For simplicity, we do not add extra noise in our main experiments. Still, this shows that it is \textit{not} important to denoise the exact noise outputted by \methodName{}; noises close to it also denoise to good actions.

\textbf{Padding noises do not affect performance.} $\pi_{0.5}$-based VLAs also predict padding action elements, such that its action expert can yield a fixed-size output~\citep{pi2025pi05}. Our reference actions use the ground-truth padding value to make actions said fixed size.
The noises corresponding to those padding elements do not seem to affect the output significantly, nor do they affect performance (as verified by setting the padding noises to $\mathcal{N}(0, I)$ after running flow reversal, then denoising). We adopt this technique in all our \methodName{} instantiations.

\subsection{Noise Averaging/Repeating is Benign or Beneficial}
\label[appendix]{app:noise-averaging}

\begin{wrapfigure}[13]{r}{0.5\linewidth}
  \centering
  \vspace{-1em}
  \includegraphics[width=\linewidth]{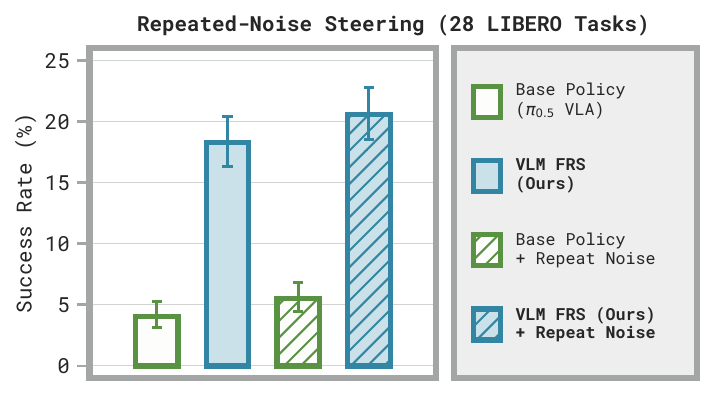}
  \caption{When running zero-shot FRS, averaging and repeating the noises used for flow reversal still yields good performance.}
  \label{fig:libero-repeat-noise-zero-shot-frs-results}
  \vspace{-5cm}
\end{wrapfigure}

The noises produced by \methodName{} tend to empirically vary across the action chunk axis less than if they were sampled i.i.d. from $\mathcal{N}(0, I)$. 
This is the case for VLM steering, where the reference action chunks $a_1$ passed through flow reversal are already repeated across the chunk axis (\cref{fig:rollout-noise-results}, center) -- though, in principle, this does not mean that the individual actions \textit{must} map to similar noises. 
It is \textit{also} the case when running \methodName{} on the LIBERO-90 data that the oracle was trained on (\cref{fig:libero90-chunk-var}) -- even though said reference actions are \textit{not} repeated, they simply may have low-frequency temporal correlations~\citep{pertsch2025fast}.

\textbf{Averaging and repeating noises.} In turn, we find that averaging and repeating the noises from \methodName{} over the chunk axis still comparably boosts performance of zero-shot \methodName{} in LIBERO, when using either VLM or oracle policy actions for steering (\cref{fig:libero-repeat-noise-zero-shot-frs-results} and \cref{fig:oracle-results} respectively). Alternatively, averaging the oracle's actions over the chunk axis still works, motivating our use of uni-directional reference actions.

This further suggests that high fidelity of flow reversal is not necessary; even when averaged, the result ends up in part of noise space mapping to a good behavioral mode. It also corroborates the findings from DSRL~\citep{wagenmaker2025dsrl}, where flow steering with repeated noises is both (1) effective at eliciting beneficial robot behaviors and (2) makes finding and predicting good noises easier. 
\textit{We therefore use noise averaging and repetition for all our LIBERO DSBC and DSRL + \methodName{} experiments.}

\subsection{Takeaways for Running \methodName{}}
\label[appendix]{app:noise-space-takeaways}

The key \methodName{} design details motivated by this analysis are:
\begin{enumerate}
    \item Running \methodName{} by averaging and repeating the noise across the chunk axis does not affect performance. All our noise policies are trained to predict such chunk-averaged noises, as it makes learning easier.
    \item Averaging and repeating the \textit{good reference actions themselves} can still find good noises. This motivates our use of directional actions for steering (rather than more sophisticated ones).
    \item Reference actions should have their padding elements set correctly. Once the underlying noise is computed with flow reversal, the padding \textit{noises} can be set to $\mathcal{N}(0, I)$ without affecting performance.
    \item When running flow reversal, there is inverse correlation between reconstruction fidelity and the resulting noises' magnitudes. Increasing the number of integration steps increases fidelity, but makes the noises higher in magnitude (i.e., more out-of-distribution). We err on the side of using fewer steps by choosing the default of 10, as (1) this makes policy inference faster, (2) we want smaller noises for easier noise policy learning, and (3) extremely high action reconstruction fidelity is not always beneficial, especially when reference actions are coarse.
\end{enumerate}
For clarity, these takeaways are also expanded upon in subsequent sections.

\section{Steering Implementation Details}
\label[appendix]{app:steering}

We now discuss the design choices we made in implementing steering interfaces for both the human operator and the VLM reasoner. We emphasize that, while empirically effective, \textbf{these systems are \textit{not} fundamental to \methodName{}, nor are they our main contribution.} More sophisticated VLM systems or human steering interfaces can likewise be used with \methodName{}. See \cref{fig:frs-overview}, left.

\subsection{Human Steering Interface}
\label[appendix]{app:human-steering-interface}

For our human \methodName{} experiments, we let an operator steer the arm via keystrokes: six keys (\texttt{W/A/S/D/Q/E}) for the three translational Cartesian axes (forward/back, left/right, up/down), a key to change gripper state, and a key to defer to the policy. Rotations are never commanded from the keyboard. 

\textbf{Extracting actions.} Each keystroke is turned into a full reference chunk of absolute joint-position targets with the DROID codebase's inverse kinematics~\citep{khazatsky2024droid}, starting from the arm's current state. Two hyperparameters govern this. First, \texttt{keyboard\_pos\_scale} ($10.0$) divides the default Cartesian step size of 1, providing the conversion between keyboard inputs and Cartesian units. This is likewise used in other past teleoperation schemes, e.g., HIL-SERL~\citep{luo2025serl, luo2025hil_serl} uses this kind of scale factor when converting a 3D SpaceMouse's inputs to Franka actions.
Second, \texttt{action\_repeats} ($5$) sets how many steps that motion is held for, i.e., the length of the reference chunk. The resulting actions are seven joint dimensions plus one gripper dimension, repeated \texttt{action\_repeats} times.

The gripper key cycles through three states: \emph{open}, \emph{closed}, or \emph{noise}. In open/closed mode the gripper dimension is driven to the commanded value and held across steering. In noise mode the gripper is left to the policy: the gripper element of the chunk is held at its current value and its corresponding noise element is randomized, so the denoised action is free to open or close based on the observation.

After passing the small reference through IK to get joint actions, we tile it until reaching the policy chunk length (15, e.g., repeat the length-5 joint angle chunks thrice). This is what gets passed through flow reversal. In turn, we set the tiled parts of the chunk (e.g., everything after the first 5 actions) to $\mathcal{N}(0, I)$ noise, allowing it to be denoised by the policy based on observation. This is akin to diffusion in-painting, albeit in noise space -- that is, we want the steered action to match part of the reference chunk, while the rest of it is ``filled in'' by what the policy thinks is reasonable.

The final human steering loop involves the person pressing a key given the current step, converting it to an action chunk programmatically using inverse kinematics, running flow reversal, reconstructing the action, then executing some or all of it before the operator is queried again.
We opt to execute the first 10 actions in the length-15 chunk. As we limit episodes to 400 low-level action steps, each episode requires the human to issue at most 40 keypress reference actions, each 10 low-level steps apart -- compared to having to provide continuous dense actions via true teleoperation.

\subsection{VLM Steering Prompt and Interface}
\label[appendix]{app:vlm-steering-prompt}

\begin{figure*}[t]
    {\tiny\texttt{\input{figures_text/VLMPrompt.txt}}}
    \caption{
        Prompt for querying Gemini for motions. Brackets indicate minor changes between LIBERO-90 and the other experiments.
    }
    \label{fig:gemini-prompt}
    
\end{figure*}

\begin{figure*}[t]
    \centering
    \includegraphics[width=\linewidth]{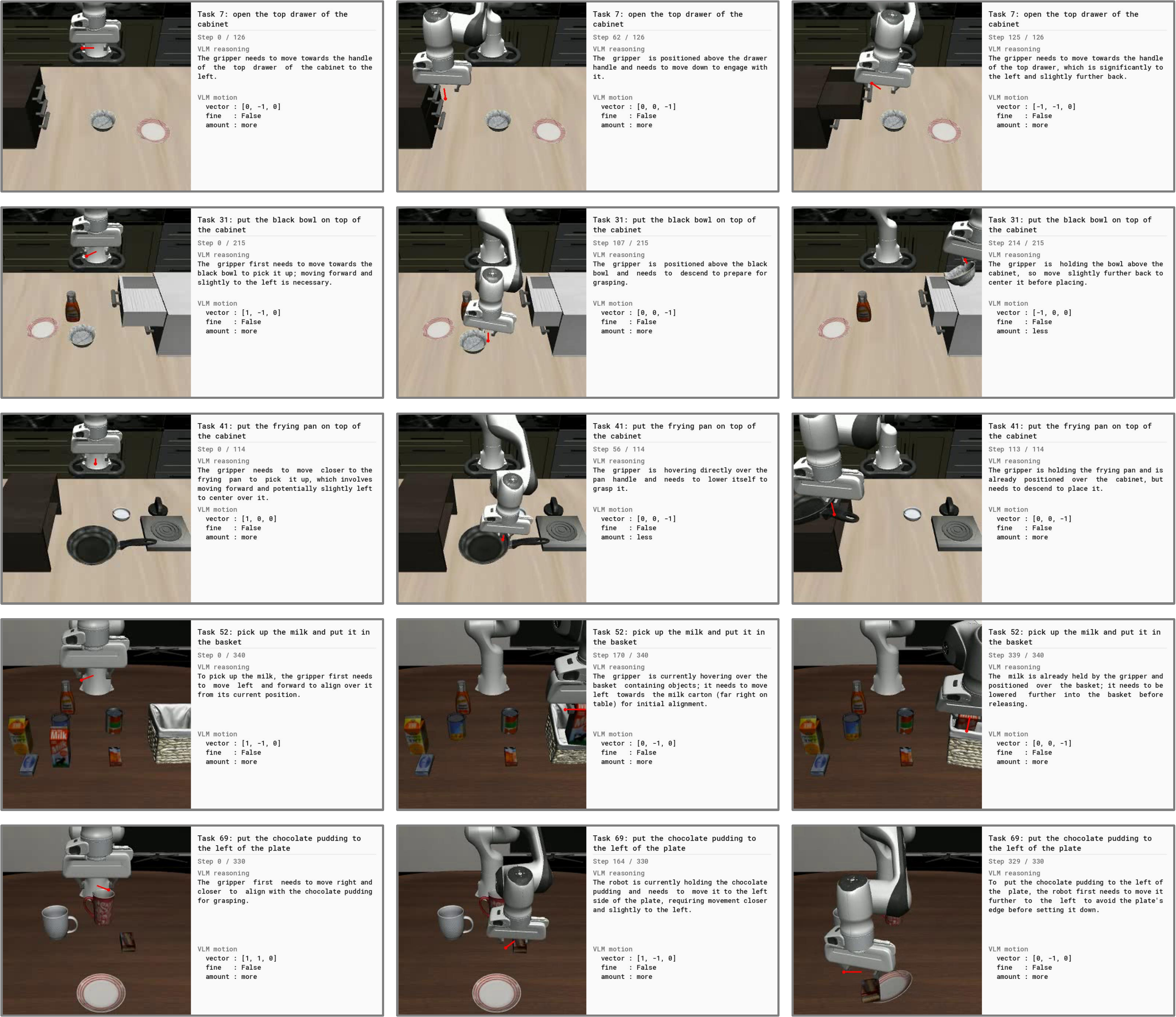}
    \caption{
        Example outputs of VLM steering in LIBERO. Red indicates the direction suggested by the VLM.
    }
    \label{fig:vlm-steering-examples}
    
\end{figure*}

For the VLM steering experiments, we use a simple prompt to infer directional actions similar to the humans'.
Specifically, we query the Gemini-ER-1.6 VLM~\citep{geminiteam2024gemini, grt2025geminirobotics} with the overall task language, and ask it produce a structured JSON output containing:
\begin{enumerate}
    \item A brief natural-language embodied chain-of-thought reasoning~\citep{zawalski2024ecot, grt2025geminirobotics, chen2026steerablePolicies}, asking it to decide the immediate task-appropriate subgoal for the robot to pursue.
    \item Whether or not the robot needs to make a fine manipulation by deferring to the base policy. If so, this overrides the below two terms.
    \item If not, the specific axis-aligned end-effector motions necessary for the robot to take, represented with a 3-element vector specifying if the motion should be positive, zero, or negative along each of the three Cartesian axes.
    \item How large said motion should be (``more'' or ``less'').
\end{enumerate}

The VLM itself receives the third-person external camera view (\textit{not} the wrist camera), which is annotated with a translucent vertical plumb line from tabletop to gripper. We include this as the front view camera used by LIBERO makes depth perception difficult; while policies are able to remediate this with wrist camera view, we found VLMs to be empirically bad at synthesizing information from third-person and wrist views (at least without substantial VLM system engineering).
We note that LIBERO provides the table height, end effector pose, and camera calibration matrices necessary to compute the pixels for the plumb line. Calibration and end effector pose are likewise easily extracted in real-world robot setups, e.g., using a ChArUco board with the DROID robot workcell~\citep{khazatsky2024droid}.

If a motion is specified in the VLM output, the vector is normalized before being projected into actions. If the motion should be small, then this vector is in turn halved. Note that this is more flexible than the human interface, as the VLM can issue \textit{combinations} of axis-aligned motions at each step.
This is a design choice -- a more sophisticated interface would allow a human to likewise issue more complex motions, albeit at the cost of intuitiveness.
See \cref{fig:gemini-prompt} for the full prompt and \cref{fig:vlm-steering-examples} for example outputs. This prompt is used for \textit{all} zero-shot VLM control and steering experiments in \cref{subsec:zero-shot-experiments}, not just \methodName{}.

Note that, as we demonstrate in \cref{app:expert-noise}, better steering actions lead to better zero-shot \methodName{} performance. While we demonstrate that \methodName{} can already achieve substantial performance gains with our simple and na\"{i}ve cardinal action VLM system, \textit{this suggests that procuring better actions from a more engineered and complex VLM system would likely improve \methodName{}'s zero-shot performance even further}. We leave such VLM engineering improvements to future work.

\textbf{Extracting actions.} LIBERO uses delta end effector actions, wherein the first three terms correspond to Cartesian motions. As $\pi_{0.5}$-LIBERO already includes normalization parameters, we consider the VLM's outputs as \textit{normalized Cartesian actions} -- that is, +1 along the x-axis means moving the maximum delta-x value recorded in the normalization statistics, -1 means moving the minimum, and 0 corresponds to the average between those two extremes (which might not be 0). 
All unnormalized rotations are set to 0 (none), as is the gripper value (halfway between fully open and closed). Note that this does \textit{not} force the end effector to never rotate, nor the gripper to stay partially open -- depending on the observed scene, both still seem to denoise to appropriate values. Alternative ways of setting these values can also work, e.g., having the VLM also set them, injecting stochasticity into those elements' noises specifically, etc.

This VLM steering system does \textit{not} require any advanced APIs for control, in contrast to past zero-shot control works~\citep{fu2026capX}; all actions it produces are with this simple programmatic conversion scheme, and are represented in the same LIBERO action space used by learned policies (as steering requires reference actions and policies' sampled actions to live in the same space).

\subsection{Action Padding} 

$\pi_{0.5}$ always predicts 32-dimensional actions, so as to accommodate predicting actions for two bimanual arms and navigation~\citep{pi2025pi05}. For DROID and LIBERO, which have 8D and 7D actions respectively, this means the remaining dimensions are filled with some padding element (-1 for DROID and 0 for LIBERO). Thus, when computing reference actions, we likewise set the padding to those values. Once the padded action has been noised via flow reversal, we find that replacing the noises corresponding to padding with true $\mathcal{N}(0, I)$ i.i.d. noise does not affect performance (\cref{app:expert-noise}). As ignoring padding noises also means noise policies need to predict from samples from a smaller output space, we universally use this technique.

\section{Simulation Experimental Details and Hyperparameters}
\label[appendix]{app:sim}

We now present details on the simulated results in \cref{subsec:zero-shot-experiments}, \cref{subsec:dsbc-experiments}, and \cref{subsec:rl-experiments}. All policy evaluations use 50 trials per task, and all error bars are Wilson confidence intervals.

\begin{table}[t]
  \centering
  \caption{Hyperparameters for DSBC and DSRL~+~FRS on LIBERO-90 with the frozen $\pi_{0.5}$ policy.}
  \label{tab:libero-hyperparams}
  \resizebox{\columnwidth}{!}{%
  \small
  \begin{tabular}{l l l l}
    \toprule
    \textbf{Hyperparameter} & \textbf{DSBC} & \textbf{DSRL~+~FRS~(\cref{app:dsrl-frs-multiple-success})} & \textbf{DSRL~+~FRS (\cref{app:dsrl-frs-one-success})}\\
    \midrule
    \multicolumn{3}{l}{\emph{Base policy \& environment}} \\
      Base policy / model family   & $\pi_{0.5}$ (\texttt{pi05\_libero}) & $\pi_{0.5}$ (\texttt{pi05\_libero}) & $\pi_{0.5}$ (\texttt{pi05\_libero}) \\
      Tanh clip magnitude     & 5.0        & 5.0  & N/A\\
      Query frequency              & 10         & 10 & 10\\
      Image resize                 & 84         & 84 & 84\\
      \midrule
      \multicolumn{3}{l}{\emph{Noise actor network}} \\
      Hidden dims                  & (128, 128, 128) & (128, 128, 128) & (128, 128, 128) \\
      Tanh-squashed policy         & True       & True & False \\
      Initial std scale            & 0.01       & 0.01  & 0.01\\
      Entropy term                 & False      & False & False\\
      \midrule
      \multicolumn{3}{l}{\emph{Optimization}} \\
      Objective                    & BC only    & SAC + BC auxiliary & SAC + BC auxiliary \\
      Training steps               & 12{,}500 & 200{,}000 & 250{,}000 \\
      Batch size                   & 128        & 16 (BC and SAC) & 16 (BC and SAC)\\
      Optimizer                    & Adam       & Adam & Adam \\
      Actor learning rate          & $1\times10^{-4}$ & $1\times10^{-4}$ & $1\times10^{-4}$ \\
      Critic learning rate         & --         & $3\times10^{-4}$ & $3\times10^{-4}$\\
      Discount $\gamma$            & --         & 0.999 & 0.999 \\
      Target smoothing $\tau$      & --         & 0.005 & 0.005\\
      Critic ensemble size         & --         & 10 & 10\\
      Critic reduction             & --         & mean & mean\\
      UTD ratio (\texttt{multi\_grad\_step}) & -- & 20 & 10 \\
      Start online updates         & --         & 1{,}000 & 1{,}000\\
      SAC actor KL penalty $\beta$ & --         & 0.005 & 0.1 \\
      \midrule
      \multicolumn{3}{l}{\emph{BC loss}} \\
      BC loss mode                 & NLL        & NLL & NLL\\
      BC main / aux coef           & 1.0 (sole objective) & 1.0 & 1.0 \\
      BC decay per online success  & None & None & 0.5 \\
      Zero-reg coef                & 0.0        & 1.0 & 0.0 \\
      Zero-reg mode                & -- & KL to $\mathcal{N}(0,I)$ & -- \\
      
      \midrule
      \multicolumn{3}{l}{\emph{FRS prefill data}} \\
      Prefill rollouts             & VLM-RL avg-noise & VLM-RL avg-noise & VLM-RL avg-noise \\
      Num.\ trajectories           & all successes        & 20 (success and failure) & 1 (success only) \\
      \bottomrule
  \end{tabular}
  }
\end{table}

\subsection{Zero-shot Experiments}
\label[appendix]{app:zero-shot-details}
\textbf{Tasks.} We run zero-shot VLM \methodName{} control on all of LIBERO-Goal, Object, and Spatial, as well as all 62 tasks that $\pi_{0.5}$-LIBERO achieves under 40\% success rate on:
\texttt{[0, 1, 3, 4, 5, 6, 7, 8, 11, 13, 14, 15, 16, 17, 18, 21, 23, 25, 26, 27, 28, 31, 32, 34, 35, 36, 37, 39, 40, 41, 42, 43, 44, 45, 48, 49, 52, 53, 58, 59, 62, 63, 64, 65, 66, 69, 71, 73, 74,75, 76, 78, 79, 80, 81, 83, 84, 85, 86, 87, 88, 89]}. All are run for 50 episodes.

\textbf{Comparisons.} We use 16 parallel action samples per step for the sample-and-rank baseline. We partially-noise to $t=0.5$ and use 5 denoising steps (each of step size $h=0.1$) for the partial noising baseline. For all LIBERO experiments, we use a chunk execution horizon of 10 (i.e., predict an action chunk of at least length 10, then execute the first 10 before re-querying the policy), as was done in past works~\citep{chen2025ecot-lite, jain2025polaris}.

\subsection{DSBC Experiments}
\label[appendix]{app:dsbc-details}

\subsubsection{Main LIBERO DSBC Details}

\textbf{Tasks.} For DSBC, we focus just on the 15 LIBERO-90 tasks where zero-shot \methodName{} makes an improvement of at least 10\% (as those are the only ones where there are enough successes to meaningfully run supervised learning on):
\texttt{13, 31, 35, 41, 43, 44, 49, 52, 69, 73, 74, 75, 78, 79, 81}.
We train on all successes of the 50 zero-shot \methodName{} rollouts per task \textit{repeated-noise} rollouts, though similar performance can be achieved with half the data or less. All comparisons are run for 50 rollouters per task.

\textbf{Hyperparameters.} All hyperparameters for DSBC are listed in \cref{tab:libero-hyperparams}.

\textbf{DSBC training.} As with all of our simulated noise policies, the DSBC policy predicts a single 7D vector, corresponding to average noises. To convert it into actual noises, we repeat it along the chunk dimension to the chunk horizon (10 for $\pi_{0.5}$-LIBERO), then pad all actions to be 32D with i.i.d. $\mathcal{N}(0, I)$ noise. As with the original DSRL implementation, we regularize the noise actions by applying tanh clipping to the noise outputs. However, we use a wider clipping range of $[-5, 5]$ (rather than $[-1. 1]$ from \citet{wagenmaker2025dsrl}), as doing so makes the noise policy more expressive, and able to fit more out-of-distribution actions' noises.

Additionally, we use negative log-likelihood (NLL) as our BC loss. This is a design choice -- we also try running with mean-square error loss, and find it works approximately just as well. This has the caveat that it does not train the policy to produce variances (Gaussian policies only learn the mean with standard MSE), and empirically benefits from regularizing the action predictions towards 0 (e.g., either with an auxiliary MSE loss that brings the mean to all 0s, or an auxiliary KL loss with $\mathcal{N}(0, I)$).

\textbf{Baselines.} For the standard BC baseline, we train a small policy of the same architecture as DSBC on the successful \textit{robot} action chunks from \methodName{} (\textit{not} noise actions). It predicts the full action chunk, which is naturally not averaged nor repeated. It is trained with NLL as the BC loss (same as our DSBC policy). This uses a Gaussian policy, which is sufficient for learning LIBERO tasks (whereas real standard BC requires a flow policy).

\subsubsection{Offline DSBC in LIBERO}
\label[appendix]{app:offline-dsbc-libero}

Online DSBC experiments is akin to HG-DAgger~\citep{kelly2019hgdagger} or online RL: it uses data from online rollouts controlled via \methodName{}, as that confirms whether or not the noises produced by noise reversal actually map to good actions (akin to receiving rewards). However, given how ``forgiving'' noise space is, can we run \methodName{} on a completely offline (normal) action dataset to train policies with DSBC?

We use the noises produced in \cref{app:ood-actions} by running \methodName{} on the entire LIBERO-90 dataset, which is usually used for standard BC. We then train individual DSBC policies on all 89 tasks\footnote{We use one the \textit{success-filtered} version of the dataset released by \citet{chen2025ecot-lite} and used for training the LIBERO-90 VLA by \citet{jain2025polaris}. Within, there was one task with 0 successes.}, including all the tasks we ran \textit{online} DSBC on. On aggregate, this boosted performance from 30\% for the base policy to 40\% with offline DSBC (whereas zero-shot VLM \methodName{} with the oracle policy achieves 50\%, see \cref{fig:oracle-results}). 

This indicates that, even if no online rollouts are used to validate whether noises from flow reversal map back to good actions, said noises can still effectively be used for DSBC. Practically, this also gives a simple recipe for turning standard demonstration data -- with normal robot actions and no noises -- \textit{into noise action datasets}. We note that DSRL can also use normal robot prior data, but requires expensive distillation of standard Q-functions into noise-action ones~\citep{wagenmaker2025dsrl}.

\subsection{DSRL + \methodName{} Experiments}
\label[appendix]{app:dsrl-frs-details}

We run two sets of DSRL + \methodName{} experiments on LIBERO-90 for a total of 25 tasks: one on the 15 tasks where \methodName{} boosts performance significantly, as well as one on a harder split of 10 tasks that both the base VLA and \methodName{} struggle on, thus allowing us to test how \methodName{} bootstraps RL in regimes where it helps zero-shot performance significantly \textit{and} when it only improves time to first success.

\subsubsection{Prefilling DSRL with Multiple \methodName{} Successes}
\label[appendix]{app:dsrl-frs-multiple-success}

\textbf{Tasks.} We first run DSRL + \methodName{} on the same 15 tasks as in DSBC. However, we do \textit{not} use all successful trajectories: instead, we sample only 20 of the 50 \methodName{} rollouts (with successes and failures proportional to how many there are in the total 50) to run BC loss on and to prefill the replay buffer. We do this over 3 seeds, each randomly sampling (potentially different) rollouts.

\textbf{Hyperparameters.} See \cref{tab:libero-hyperparams}.

\textbf{DSRL + \methodName{} training.} Prefilling replay buffers is an effective way to consume prior data in RL~\citep{ball2023rlpd}. As \methodName{} produces expert noise actions during zero-shot steering, it allows DSRL to likewise use those trajectories as prior data for policy and critic learning, without any changes -- given that DSRL uses SAC~\citep{haarnoja2018sac-softActorCritic} to learn the noise policy, it can consume off-policy data.

The only change we make to the actual RL algorithm is adding a BC loss term to actor learning:
\begin{equation}
    \mathcal{L}_\text{DSRL + FRS, actor} = \mathcal{L}_\text{DSRL, actor} + \beta \mathcal{L}_\text{BC}
\end{equation}
This is a popular soft behavior constraint~\citep{luo2023finetuningOfflineToOnlineRL, fujimoto2021minimalistApproachOfflineRL}, especially when used with SAC. As with DSBC, we use NLL as the BC loss, but find that using MSE loss achieves similar performance, especially in conjunction with a 0-action regularizer. We use BC loss weight $\beta = 1$.

At each learning step, two batches of equal batch size are sampled: one for standard SAC learning, the other for the BC regularization loss (which is drawn exclusively from the successful prefill trajectories).

Our noise policy is parameterized in the same way as the DSBC one: it predicts average noises (which are repeated and padded during inference), and it uses tanh clipping.

\textbf{Baselines.} For the DSRL baseline, we use the exact same hyperparameters as DSRL + \methodName{}. This includes predicting a single noise corresponding to all non-padding actions, then repeating it across the chunk length and adding unit Gaussian padding for the remaining dimensions.

For the residual RL comparison, we follow some prescriptions from PLD~\citep{xiao2025pld-probeLearnDistill}, as they also apply residual RL to VLAs. We control for policy architecture with respect to DSRL. To determine the residual RL maximum edit distance, we use $0.5$ as the tanh clipping coefficient, as that is the distance used in PLD (which the authors mention being sensitive to tuning). However, as we are \textit{not} aiming to distill the residual policy's outputs into the VLA, we simplify our residual RL implementation from PLD by removing the ``probing'' stage (wherein the residual policy defers to the base VLA for some number of steps), as this mainly serves to help the policy act as an expert from more starting states. Our residual policy edits the full chunk of the base VLA, and is otherwise trained with the same SAC~\citep{haarnoja2018sac-softActorCritic} implementation and hyperparameters as our DSRL runs.

Additionally, the data buffers are filled the exact same way as in DSRL (by running online policy rollouts until 1000 frames are collect). Critically, we also do not use PLD's pretraining phase, as doing so requires collecting 50 successful rollouts per task \textit{from the base VLA}. \textit{Given that we are targeting tasks where the base VLA only achieves 0 or 1 successes in 50 rollouts, collecting 50 successes would be impractical.}

\subsubsection{Prefilling DSRL with One \methodName{} Success}
\label[appendix]{app:dsrl-frs-one-success}

\textbf{Tasks.} We consider 10 LIBERO-90 tasks that the base VLA get close to 0\% average success rate on \textit{and} which zero-shot \methodName{} achieves an average performance of under 10\% on (vs. over 30\% on the previous 15-task split). The task IDs are:
\texttt{1, 3, 18, 25, 36, 37, 49, 63, 69, 78}. The DSRL and DSRL + \methodName{} baselines are run for three seeds, RoboMeter only uses one.

\textbf{KL-based DSRL.} In cases with more difficult and low-success tasks, DSRL needs to be able to sample (rarely) more out-of-distribution actions compared to the vanilla clipped DSRL, which can only sample within $[-1, 1]$ due to a tanh activation. Therefore, we switch to not using the tanh activation after predicting a mean and standard deviation for a Gaussian, allowing the policy to express all means. To prevent the policy from exploiting this, we add a regularization term of $0.1$ KL relative to the standard normal distribution $N(0, 1)$ which incentivizes the policy to stay close to the original DSRL distribution. We find that without this change, overall DSRL performance is very low.

\textbf{Buffer filling.} Since DSRL + \methodName{} may have only one or two successes within 50 rollouts on difficult tasks, we instead run zero-shot VLM \methodName{} until we get one success, and then only load that single success into the replay buffer to jump-start reinforcement learning. 

\textbf{Hyperparameters.} See \cref{tab:libero-hyperparams}.

\textbf{Baselines.} For robometer, see the dedicated section for VLM reward methods in \cref{app:robometer}. For DSRL vanilla, we use the same formulation as KL-DSRL without the warmstart.

\subsection{VLM Reward Methods}
\label[appendix]{app:robometer}

For VLM reward methods, most VLM reward methods also train \textit{directly} on LIBERO, therefore containing privileged information about success trajectories that we do not assume in our DSRL+FRS approach. We compare with Robometer, a work which is trained generally on a broad set of simulation and real-world environments. We modify the base SAC reward to have reward \texttt{-(1 - dense-reward)} instead of the default \texttt{-1} for each non-terminal step, and \texttt{0} in each terminal step, in the same way as \citep{liang2026robometer}. We do not use the transformer-based chunk prediction modification to DSRL stated in the aforementioned paper. We find that our method greatly outperforms Robometer-guided DSRL, possibly because FRS more effectively focuses the policy onto good action modes without relying on the base policy already choosing good actions at a substantial rate. 

\section{Real-World Details and Hyperparameters}
\label[appendix]{app:real-world}

We run our real-world experiments on the DROID Franka setup~\citep{khazatsky2024droid}, using the open-source $\pi_{0.5}$-DROID VLA~\citep{pi2025pi05}. All evaluations use 20 rollouts per task, and all error bars are Wilson confidence intervals.

\textbf{Tasks.} We consider six tasks, shown in \cref{fig:real-results}. All involve leveraging skills captured within the DROID dataset, albeit in ways or situations that are uncommon or novel.

\textbf{Zero-shot human \methodName{}.} We run zero-shot human-in-the-loop \methodName{} using the interface from \cref{app:human-steering-interface}. 
The operator attempts 20 trials of each of the six tasks, recording successful ones.
Each episode, the human can issue one task per chunk.
As we use receding horizon control (wherein a full action chunk is predicted, the first 10 are executed, and then the policy is queried again) and maximum episode length 400, the person has issue only 40 keystrokes per episode at most.

\textbf{Real-world DSBC.} DSBC reuses 10 successes from the DROID zero-shot \methodName{} experiments (of which all tasks achieve). All frames from these tasks are used for supervised learning with DSBC, amounting to at most 400 frames.
We parameterize our DSBC policy in the same way as \citet{wagenmaker2025dsrl} (albeit without attending into the flow VLA). However, it does \textit{not} predict averaged and repeated noises, instead predicting the full chunk of noises (no padding, as usual). Additionally, we use MSE loss (and thus do not learn variance) for all tasks except the towel hanging task (see below).

\textbf{Real-world DSRL + \methodName{}.} We run a simple form of DSRL + \methodName{} on the towel hanging task (which was empirically the most difficult one). Specifically, we use batched policy gradients, where all successful trajectories' actions have reward 1 and all unsuccessful have -0.1. This requires action probabilities, motivating us to train on this task with NLL.
After training the original BC policy, we roll it out 10 times, then update the policy with 2.5k steps on that data. This is repeated once (i.e., two batches of online rollout updates). The final policy is likewise evaluated over 20 rollouts.
All hyperparameters are shared with DSBC, albeit with fewer gradient steps.

\textbf{Hyperparameters.} See \cref{tab:realworld-bc-hyperparams}.

\textbf{Baselines.} We compare against the base policy and a standard BC baseline. The latter is trained on the same data as the DSBC policies (i.e., from the steered rollouts), but are trained to output a chunk of \textit{normal robot actions}, not noises. 

We initially tried both a simple Gaussian policy and an expressive flow-matching one.
They use the same overall architecture as DSBC (of course, with the final layer training on different objectives, e.g., flow matching). 
While we ensured both implementations were correct by training on a simpler DROID pick-and-place task, both failed at the six tasks that DSBC excelled at.
In particular, DSBC's ability to ``fall back on'' the VLA's action prior made it much more data efficient, allowing it to learn these challenging tasks in just 10 trajectories.
However, as flow BC achieved non-zero success rate for the ``hang tape on black stand'' task (which we use for real-world offline DSBC), \textbf{we consider the flow policy the definitive standard BC approach for our real-world experiments.}

Additionally, we found it empirically important to output normalized actions and input proprioceptive states, so we use the same open-source DROID normalization statistics as $\pi_{0.5}$-DROID for this.
Both standard BC policies we tried use the same hyperparameters, though we just train for 4x as long for the flow policies (as flow and diffusion tend to converge slower).

\begin{table}[t]
    \centering
    \caption{Hyperparameters for real world learning. 
    DSBC and standard BC with a Gaussian policy share hyperparameters.
    All share the same dual-camera DSRL-style encoder and MLP
    trunk; they differ in \emph{what} they predict (the VLA's flow reversal noise
    vs.\ the robot actions directly) and in the output head (Gaussian
    regression vs.\ flow matching). The DSBC head is trained with MSE by default; an
    NLL (Gaussian, mean + log-std clamped to $[-5,2]$) variant is used only to initialize
    RL finetuning.}
    \label{tab:realworld-bc-hyperparams}
    \resizebox{\columnwidth}{!}{
    \small
    \begin{tabular}{l l l}
      \toprule
      \textbf{Hyperparameter} & \textbf{DSBC} & \textbf{Flow-matching Standard BC} \\
      \midrule
      \multicolumn{3}{l}{\emph{Task setup \& I/O}} \\
        Base policy            & $\pi_{0.5}$ (\texttt{pi05\_droid\_jointpos}) & none (standalone) \\
        Image resize                        & $84\times84$ & $84\times84$ \\
        Action chunk length                 & 15 & 15 \\
        Query frequency (open-loop horizon) & 10 & 10 \\
        Prediction target                   & Flow-reversal noise & robot actions \\
        Per-step output dim                 & 15 x 8 (noise) & 15 x 8 (action) \\
        Proprioception input                & No & Yes \\
      \midrule
      \multicolumn{3}{l}{\emph{Policy network (shared trunk)}} \\
        Image encoder (per camera)          & DSRL small CNN, 4 conv, stride $2/1/1/1$ & DSRL small CNN, 4 conv, stride $2/1/1/1$ \\
        Latent bottleneck (per camera)      & Linear$\to$LayerNorm$\to$tanh, dim 50 & Linear$\to$LayerNorm$\to$tanh, dim 50 \\
        MLP hidden dims                     & (256, 256, 256) & (256, 256, 256) \\
        Output head                         & deterministic mean & flow velocity field $v_\theta(x_t,t,\mathrm{obs})$ \\
        Timestep embedding                  & -- & sinusoidal (sin/cos), dim 64 \\
      \midrule
      \multicolumn{3}{l}{\emph{Optimization}} \\
        BC loss                             & MSE & conditional flow matching \\
        Training steps                      & 12{,}500 & 50{,}000 \\
        Batch size                          & 128 & 128 \\
        Optimizer                           & Adam & Adam \\
        Learning rate                       & $3\times10^{-4}$ & $3\times10^{-4}$ \\
      \midrule
      \multicolumn{3}{l}{\emph{Flow matching}} \\
        Flow-time sampling                  & -- & $\mathrm{Beta}(1.5,1)\cdot 0.999 + 0.001$ \\
        ODE solver (inference)              & -- & Euler, 10 steps ($t{=}1{\to}0$) \\
        Action normalization                & none ($\epsilon\sim\mathcal{N}(0,I)$ already) & per-dim $z$-score (data stats), un-normalized at sampling \\
      \midrule
      \multicolumn{3}{l}{\emph{Training data}} \\
        Trajectories                        & successful only & successful only \\
        Num.\ trajectories per task         & 10 & 10 \\
      \bottomrule
    \end{tabular}
    }
  \end{table}

%% file: figures_text/VLMPrompt.txt
You are an expert giving commands to a robot gripper from a third-person view. Focus on the gripper (centered at the top of the blue line), not the base of the arm. The blue line coming down from the gripper ends at the table surface (but is NOT an object, only a guide for navigation).

\smallskip

The task is: <task language>

The task language refers to locations from the camera's perspective. We should use the following steps to get to our solution:

1. Understand the task: left and right are from the image's perspective. Front is closer to the foreground, back is closer to the background and the robot arm's base. Up is away from the table, down is closer to it.

2a. Output one line describing the immediate next steps to accomplish this task. If we are still far from our next destination, we should say how we should move spatially to get there. If we are close, describe whether smaller (fine) motor adjustments are necessary or if spatial movement is sufficient. If manipulation may be in progress (gripper not closed around an object and not open and away from object) default towards fine motor skills.

2b. Use the bottom of the blue line (on the table surface) to judge whether the gripper needs to move closer (bring bottom of line towards bottom of image) or farther (bottom of line towards top of image) from the camera. In these situations, default to a zero z-axis command unless the gripper is roughly aligned with the object or needs to move up to avoid obstruction.

3. Output our structured action: Set the "fine" flag if fine motor skills (e.g. positioning, grasping, pushing, pulling, rotating) are needed as the next low-level action. Otherwise, give the principal axes of the next spatial motion in [x, y, z] space. If we could potentially get stuck on nearby obstacles, prefer to move in the x-y plane before moving down and prefer to move up before moving in the x-y plane.

4. Use the "motion\_amount" field to determine how much to move. E.g., if the gripper is far from the target object, "motion\_amount" should be "more" to indicate that significant motion in the specified direction is needed. If the gripper is close to the target object, "motion\_amount" should be "less" to indicate that only a small adjustment in the specified direction is needed.

\smallskip

Hints:

[Base:

- The ketchup is in a bottle, while the tomato sauce is in a red can.

]

[Transfer:

- The alphabet soup is in a blue can, while the tomato sauce is in a red can.

- The ramekin is white, the cookie box is mainly yellow and red.

- The salad dressing is in a white and green bottle. The ketchup is in a red bottle with a white lid. The BBQ sauce is in a small red bottle.

- The butter is in a red box, while the cream cheese is in a blue box.

]

- "Middle" does not necessarily mean centered in the table, it means relative to other objects. E.g., if there are three bowls, then the middle bowl is the one that is between the other two, even if it's not in the middle of the scene.

\smallskip

The output should be a JSON with fields:

"reasoning": str, brief 1 sentence reasoning following the steps indicated above.

"fine": bool.

"coords": [x, y, z] with each of x, y, z in -1, 0, 1.

"motion\_amount": str, either "less" or "more".

\smallskip

fine:

False = coarse motion: use `coords` to steer. Set `coords` to the discrete direction the gripper should move next; combine axes when motion is oblique.

True = fine / manipulate (close fingers on object, place, hold steady, fine vertical adjustments). When `fine = True`, `coords` should be `[0, 0, 0]`.

\smallskip

Axes for coords:
x = **farther** from camera (-1) / none (0) / **closer** to camera (+1)
y = image **left** (-1) / none (0) / image **right** (+1)
z = world **down** (-1) / none (0) / world **up** (+1)

\smallskip

`motion\_amount`:

"less" = Less motion in the specified direction is needed.

"more" = Significant motion in the specified direction is needed.

\smallskip

If you are close to grasping or placing an object, ALWAYS prefer outputting fine manipulation over coords and less motion! Motion should only be used for larger adjustments.